\documentclass[11pt]{article}

\usepackage[final]{acl}

\usepackage{times}
\usepackage{latexsym}

\usepackage[T1]{fontenc}

\usepackage[utf8]{inputenc}

\usepackage{microtype}

\usepackage{inconsolata}

\usepackage{graphicx}

\usepackage{hyperref}
\usepackage{booktabs}
\usepackage{multirow}
\usepackage{xcolor}
\usepackage{amsmath}
\usepackage[dvipsnames]{xcolor}
\usepackage{subcaption}
\usepackage{lipsum}

\usepackage{pifont} %
\usepackage{amssymb} %
\usepackage{siunitx}
\usepackage{xfp}
\usepackage{bbm}

\usepackage[most]{tcolorbox}
\usepackage{enumitem}
\usepackage{mathtools}

\usepackage[normalem]{ulem}

\definecolor{tabblue}{HTML}{1F77B4}     %
\definecolor{taborange}{HTML}{FF7F0E}   %
\definecolor{tabgreen}{HTML}{2CA02C}    %
\definecolor{chosenColor}{HTML}{2B5C8F} %
\definecolor{rejectedColor}{HTML}{D95F02} %

\definecolor{posgreen}{HTML}{1B8A3A}
\definecolor{negred}{HTML}{D9381E}
\definecolor{zerogray}{HTML}{7F8C8D}

\newcommand{\pos}[1]{\textcolor{posgreen}{+#1}}
\newcommand{\negval}[1]{\textcolor{negred}{#1}}
\newcommand{\zero}[1]{\textcolor{zerogray}{#1}}

\newcommand{\olmoDataset}{\texttt{\href{https://huggingface.co/datasets/allenai/olmo-2-0425-1b-preference-mix}{olmo2-1b-prefmix}} }
\newcommand{\olmoDatasetDeltaQwenSmall}{\texttt{olmo2-1b-prefmix-delta} }

\newcommand{\olmoTwoSft}{\textsc{\href{https://huggingface.co/allenai/OLMo-2-0425-1B-SFT}{OLMo2-1B-SFT}} }
\newcommand{\mistral}{\textsc{\href{https://huggingface.co/mistralai/Mistral-7B-Instruct-v0.2}{Mistral-7B-Instruct-v0.2}} }
\newcommand{\mistralPlot}{\textsc{\href{https://huggingface.co/mistralai/Mistral-7B-Instruct-v0.2}{Mistral-7B-Instruct}} }
\newcommand{\qwenTwoSmall}{\textsc{\href{https://huggingface.co/Qwen/Qwen2-1.5B-Instruct}{Qwen2-1.5B-Instruct}} }
\newcommand{\qwenOneHalfSmall}{\textsc{\href{https://huggingface.co/Qwen/Qwen1.5-4B-Chat}{Qwen1.5-4B-Chat}} }
\newcommand{\qwenOneHalfBig}{\textsc{\href{https://huggingface.co/Qwen/Qwen1.5-7B-Chat}{Qwen1.5-7B-Chat}} }
\newcommand{\aya}{\textsc{\href{https://huggingface.co/CohereLabs/aya-23-8B}{Aya-23-8B}} }
\newcommand{\gemmaTwoTiny}{\textsc{\href{https://huggingface.co/google/gemma-2-2b-it}{Gemma2-2B-it}} }

\newcommand{\qwenThreeTiny}{\textsc{\href{https://huggingface.co/Qwen/Qwen3-0.6B}{Qwen3-0.6B}} }
\newcommand{\qwenThreeSmall}{\textsc{\href{https://huggingface.co/Qwen/Qwen3-4B}{Qwen3-4B}} }
\newcommand{\qwenThreeBig}{\textsc{\href{https://huggingface.co/Qwen/Qwen3-32B}{Qwen3-32B}} }
\newcommand{\olmoTwoTiny}{\textsc{\href{https://huggingface.co/allenai/OLMo-2-0425-1B-Instruct}{OLMo2-1B-Instruct}} }
\newcommand{\olmoTwoSmall}{\textsc{\href{https://huggingface.co/allenai/OLMo-2-1124-7B-Instruct}{OLMo2-7B-Instruct}} }
\newcommand{\olmoTwoBig}{\textsc{\href{https://huggingface.co/allenai/OLMo-2-0325-32B-Instruct}{OLMo2-32B-Instruct}} }
\newcommand{\gemmaTwoSmall}{\textsc{\href{https://huggingface.co/google/gemma-2-9b-it}{Gemma2-9B-it}} }
\newcommand{\gemmaTwoBig}{\textsc{\href{https://huggingface.co/google/gemma-2-27b-it}{Gemma2-27B-it}} }

\newcommand{\qwenTwoHalf}{\textsc{\href{https://huggingface.co/Qwen/Qwen2.5-7B-Instruct}{Qwen2.5-7B-Instruct}} }
\newcommand{\tuluTwo}{\textsc{\href{https://huggingface.co/allenai/tulu-2-7b}{Tulu-2-7B}} }
\newcommand{\qwenTwoTiny}{\textsc{\href{https://huggingface.co/Qwen/Qwen2-0.5B-Instruct}{Qwen2-0.5B-Instruct}} }

\newcommand{\yi}{\textsc{\href{https://huggingface.co/01-ai/Yi-6B-Chat}{Yi-6B-Chat}} }
\newcommand{\smolLMTwo}{\textsc{\href{https://huggingface.co/HuggingFaceTB/SmolLM2-1.7B-Instruct}{SmolLM2-1.7B-Instruct}} }

\newcommand{\qwenThreeBigTiny}{\textsc{Qwen3/\href{https://huggingface.co/Qwen/Qwen3-32B}{32}-\href{https://huggingface.co/Qwen/Qwen3-0.6B}{0.6}B} }
\newcommand{\qwenThreeSmallTiny}{\textsc{Qwen3/\href{https://huggingface.co/Qwen/Qwen3-4B}{4}-\href{https://huggingface.co/Qwen/Qwen3-0.6B}{0.6}B} }
\newcommand{\olmoTwoBigTiny}{\textsc{Olmo2-it/\href{https://huggingface.co/allenai/OLMo-2-0325-32B-Instruct}{32}-\href{https://huggingface.co/allenai/OLMo-2-0425-1B-Instruct}{1}B} }
\newcommand{\olmoTwoSmallTiny}{\textsc{Olmo2-it/\href{https://huggingface.co/allenai/OLMo-2-1124-7B-Instruct}{7}-\href{https://huggingface.co/allenai/OLMo-2-0425-1B-Instruct}{1}B} }
\newcommand{\gemmaTwoBigSmall}{\textsc{Gemma2-it/\href{https://huggingface.co/google/gemma-2-27b-it}{27}-\href{https://huggingface.co/google/gemma-2-9b-it}{9}B} }
\newcommand{\avgPairs}{\textsc{Average Pairs} }
\newcommand{\fewPairs}{\textsc{Few Pairs} }
\newcommand{\morePairs}{\textsc{More Pairs} }

\newcommand{\fmtnum}[2][]{%
  #1\num{#2}%
}

\newcommand{\numColor}[2][]{%
  \ifdim \fpeval{#2}pt > 0pt
    \pos{\fmtnum[#1]{#2}\%}%
  \else\ifdim \fpeval{#2}pt < 0pt
    \negval{\fmtnum[#1]{#2}\%}%
  \else
    \zero{\fmtnum[#1]{#2}\%}%
  \fi\fi
}

\newcommand{\avg}[7]{%
  \fpeval{(#1+#2+#3+#4+#5+#6+#7)/7}%
}

\newcommand{\row}[8]{%
  #1 &
  \fmtnum{#2}\% & \fmtnum{#3}\% & \fmtnum{#4}\% & 
  \fmtnum{#5}\% & \fmtnum{#6}\% & \fmtnum{#7}\% & \fmtnum{#8}\% &
  \fmtnum{\avg{#2}{#3}{#4}{#5}{#6}{#7}{#8}}\% \\
}

\newcommand{\rowColor}[8]{%
  #1 &
  \numColor{#2} & \numColor{#3} & \numColor{#4} &
  \numColor{#5} & \numColor{#6} & \numColor{#7} & \numColor{#8} &
  \numColor[\bfseries]{\avg{#2}{#3}{#4}{#5}{#6}{#7}{#8}} \\
}

\title{Exploring the Limits of the Delta Learning Hypothesis: Single-Pair Generator Biases in Direct Preference Optimization}

\author{
Giuseppe Curci$^{1}$ \quad
Stefano Simonazzi$^{2}$ \quad
Andrea Molinari$^{3}$ \quad
Andrea Zugarini$^{2}$ \\[0.5em]
$^{1}$Okkam, Italy\quad
$^{2}$Villanova.ai, Italy\quad
$^{3}$University of Trento, Italy \\[0.5em]
\texttt{curci@okkam.it,  andrea.molinari@unitn.it,}\\
\texttt{\{ssimonazzi, azugarini\}@villanova.ai}
}

\begin{document}

\maketitle

\begin{abstract}
Direct Preference Optimization (DPO) typically relies on preference pairs generated from diverse model pools and ranked by an external LLM judge. The recently proposed delta-learning hypothesis suggests that comparable supervision can instead be obtained from a single pair of models by treating the larger model’s response as preferred. We investigate the limits of this approach across model families and scales. We show, both theoretically and empirically, that fixed model pairs introduce systematic, generator-dependent biases that are predictive of the preference label independent of response quality. Consequently, DPO trained on single-pair data consistently underperforms multi-model preference data, despite the larger generator producing higher-quality responses, and we trace this gap to training dynamics consistent with reward hacking on these biases. We further show that simply increasing the number of generators, even without an external judge, is enough to partially mitigate this effect. Our results highlight generator diversity as a critical factor in preference-data construction.

\end{abstract}

\section{Introduction}

Direct Preference Optimization (DPO)~\citep{dpo} is a widely adopted post-training technique for improving large language models (LLMs) across various tasks~\citep{dpo_safety,zhang-etal-2025-codedpo}. It remains highly relevant because it is computationally cheaper than on-policy learning, highly effective in non-verifiable domains, and provides a stronger foundational model than Supervised Fine-Tuning (SFT) alone for subsequent Reinforcement Learning with Verifiable Rewards (RLVR)~\cite{olmo2026olmo3}.

DPO requires a dataset of prompts, each paired with two completions of differing quality designated as ``chosen'' and ``rejected''. The training process then optimizes the model to assign a higher probability to the chosen completion. Often, these completions are generated by other LLMs, and their relative quality is evaluated using an LLM-as-a-judge~\cite{olmo20252olmo2furious, ultrafeedback}. 
However, this evaluation method is imperfect; judge preferences are often swayed by superficial factors like length~\citep{llm-judge-length-bias}, formatting~\citep{format_bias}, or judge-generator distributional similarity~\citep{pref_leakage}, introducing significant ranking instability~\citep{zhao2026jaggedjudgesepistemicstability}. 
Given the critical need for high-quality data, extensive research has focused on mitigating these biases~\cite{length-bias-correction}, developing data-aware training algorithms~\citep{beta_dpo}, refining data cleaning techniques~\citep{clean_first, principled_data_selection}, and evaluating the alignment between data pairs and trained models~\citep{valuable_pref_data}.

As an alternative, \citet{delta_learning_hyp} proposed the ``Delta Learning hypothesis,'' suggesting that ranking can be determined using just two models without external supervision. Thus, preference pairs are generated directly from a single model pair within the same family and lineage: a larger model for the chosen and a smaller one for the rejected. Contrary to \citet{what_matters_data_dpo}, they show how the chosen model doesn't need to be stronger than the trained model, provided there is a sufficient quality delta between the completions and \textit{minimal noise}. This method often aligns with judge preferences and drives comparable improvements. The Olmo3 developers also utilized this approach, demonstrating that LLM-as-a-judge data and the delta heuristic provide complementary signals rather than one being strictly superior~\cite{olmo2026olmo3}. Furthermore, \citet{decompose_dpo_delta_learning_cot} expanded on this by investigating the single-pair method's impact on reasoning models and Chain-of-Thought (CoT) improvements. Despite these findings, little is known about how this hypothesis generalizes across different model families and scales. A critical open question remains: \textbf{is the noise that can overwhelm the delta learning signal merely occasional correctness errors, as originally assumed, or can it take a more systematic, generator-dependent form that persists regardless of scale or dataset size?}

In this work, we present a comprehensive theoretical and empirical study comparing the two approaches. We theoretically prove that relying on a single model pair increases reward hacking by exploiting generator-dependent spurious features, and empirically confirm this mechanism through both the reward-accuracy gap in training dynamics and the classification accuracy achieved by a classifier observing only style features of chosen and rejected. Consistent with this analysis, we further demonstrate that single-pair DPO systematically underperforms relative to using multiple pairs. Finally, we show that increasing generator diversity alone, without an external judge, partially mitigates this effect for most models tested, providing a better judge-free baseline for constructing preference pairs from a model pool.

\section{Methodology}

\subsection{Direct Preference Optimization}

Let $x$ denote a prompt, and let $y_c$ and $y_r$ represent the chosen and rejected completions, respectively, sampled from a preference dataset $\mathcal{D}$. Given a policy model $\pi_\theta(y|x)$ and a reference model $\pi_{\text{ref}}(y|x)$, the length-normalized DPO loss is defined as:

\begin{equation}
\label{eq:dpo_loss_len_norm}
\log \sigma \left( \frac{\beta}{|y_c|} \log \frac{\pi_\theta(y_c|x)}{\pi_{\text{ref}}(y_c|x)} -  \frac{\beta}{|y_r|} \log \frac{\pi_\theta(y_r|x)}{\pi_{\text{ref}}(y_r|x)} \right)
\end{equation}

where $|y|$ represents the token length of completion $y$, $\beta$ is a scaling hyperparameter, and $\sigma(\cdot)$ is the logistic sigmoid function. 

\subsection{Delta Learning Hypothesis Limits}
\label{sec:delta_learning_limits}

\paragraph{Delta learning hypothesis.} The delta learning hypothesis holds that training succeeds when a stronger teacher $\theta_c$ is better aligned with the optimal parameters $\theta^*$ than a weaker teacher $\theta_r$, provided the correctness signal is not overwhelmed by spurious noise. We extend this notion of noise beyond correctness errors: relying on only two generators risks injecting \emph{systemic, dataset-wide} biases into the preference data, degrading training stability and downstream performance even when $\theta_c \succ \theta_r$ holds exactly.

\paragraph{Features.} Let $\mathcal{Y}$ be the space of completions, and let a \emph{feature} be a function $f : \mathcal{Y} \rightarrow \{0,1\}$ indicating the presence of some property in a completion (e.g., use of bullet points, correctness). We partition the features relevant to DPO into two disjoint sets: $\mathcal{F}_l$, features we want the policy to learn to prefer (correctness, safety, helpfulness), and $\mathcal{F}_s$, \emph{spurious} features we want the policy to remain indifferent to (formatting, length, style markers).

\paragraph{Desideratum.} For a preference dataset $\mathcal{D}$ to isolate signal in $\mathcal{F}_l$, no spurious feature should be a useful predictor of the preference label. We formalize this via the accuracy achievable by a classifier that only observes $f$:
\begin{equation}
\label{eq:acc_f}
    \mathrm{Acc}_f \;\coloneqq\; \mathbb{P}\big(f(y_c) > f(y_r)\big) + \tfrac{1}{2}\,\mathbb{P}\big(f(y_c) = f(y_r)\big),
\end{equation}
the probability that thresholding on $f$ alone recovers the true chosen/rejected ordering (ties broken at chance). Our desideratum is that every spurious feature be statistically uninformative:
\begin{equation}
\label{eq:independent_spurious_feat}
    \mathrm{Acc}_f = \tfrac12 \qquad \text{for all } f \in \mathcal{F}_s.
\end{equation}
Equation~\ref{eq:independent_spurious_feat} does not require that $f$ be absent from either completion, but only that its presence carry no systematic information about which completion was preferred. Note that $\mathrm{Acc}_f$ is defined with respect to the fixed orientation "$f(y_c) > f(y_r)$"; since every feature $f$ has a complementary feature (e.g. "use of \#" vs "no use of \#") $1-f$ with $\mathrm{Acc}_{1-f} = 1-\mathrm{Acc}_f$, we adopt the convention of reporting $\mathrm{Acc}_f$ for whichever orientation satisfies $\mathrm{Acc}_f \geq \tfrac12$, so that $\mathrm{Acc}_f$ coincides with the accuracy of the best fixed-direction classifier using $f$.

\paragraph{Assumption 1 (Independent generation).} \phantomsection \label{assumption:ind_generation} Given a prompt $x$ and a pair of generators $(\theta_c, \theta_r)$, completions are sampled independently, $y_c \sim \pi_{\theta_c}(\cdot \mid x)$ and $y_r \sim \pi_{\theta_r}(\cdot \mid x)$, with no significant shared randomness. For a generator $\theta$ and feature $f$, define its \emph{propensity} $p_f(\theta) \coloneqq \mathbb{E}_{x, \, y \sim \pi_\theta(\cdot\mid x)}[f(y)]$, i.e., the base rate at which $\theta$'s completions exhibit $f$.

\paragraph{Proposition 1 (Fixed-pair sampling induces systematic bias).}
\phantomsection\label{prop:fixed_pair_bias}
\emph{Let $\mathcal{D}$ be constructed from a single fixed pair $(\theta_c, \theta_r)$ for every prompt. Then for any $f \in \mathcal{F}_s$,}
\begin{equation}
\label{eq:prop1}
    \mathrm{Acc}_f = \tfrac12 + \tfrac12\big\vert p_f(\theta_c) - p_f(\theta_r)\big\vert.
\end{equation}

\emph{Proof.} Write $p_c = p_f(\theta_c)$, $p_r = p_f(\theta_r)$. By \hyperref[assumption:ind_generation]{Assumption 1}, $f(y_c)$ and $f(y_r)$ are independent Bernoulli variables with means $p_c$ and $p_r$. Then
\begin{equation}
\label{proof:fixed_pair_induces_bias}
\begin{aligned}
    \mathrm{Acc}_f 
    &= \mathbb{P}(f(y_c){=}1, f(y_r){=}0) \\
    &\qquad + \tfrac12\Big[\mathbb{P}(f(y_c){=}1,f(y_r){=}1) \\
    &\qquad + \mathbb{P}(f(y_c){=}0,f(y_r){=}0)\Big] \\
    &= p_c(1-p_r) + \tfrac12\big[p_c p_r + (1-p_c)(1-p_r)\big] \\
    &= \tfrac12 + \tfrac12(p_c - p_r)
\end{aligned}
\end{equation}

Equation~\ref{eq:prop1} already shows a direct, monotonic relationship between the two generators' propensity gap and classifier accuracy, with $\mathrm{Acc}_f \to 1$ as $p_c \to 1, p_r \to 0$. As stated in \hyperref[assumption:ind_generation]{Assumption 1}, however, $p_f(\theta)$ marginalizes over the entire prompt distribution. The corollary below shows the bias is in fact stronger than this dataset-wide average suggests when we account for the prompt distribution.

\paragraph{Corollary 1 (Domain-conditional bias).}
\phantomsection\label{cor:domain_conditional_bias}
\emph{Let $D \subseteq \mathcal{X}$ be any well-defined subset of prompts (a ``domain,'' e.g., math problems or coding tasks). Define the domain-conditional propensity $p_f(\theta \mid D) \coloneqq \mathbb{E}_{x \in D,\, y \sim \pi_\theta(\cdot|x)}[f(y)]$ and propensity gap $\Delta_f(\theta_c,\theta_r \mid D) \coloneqq p_f(\theta_c\mid D) - p_f(\theta_r \mid D)$. Under \hyperref[assumption:ind_generation]{Assumption 1} restricted to $x \in D$,}
\begin{equation}
\label{eq:corollary_domain}
    \mathrm{Acc}_f \mid D = \tfrac12 + \tfrac12\,\left\vert\Delta_f(\theta_c,\theta_r \mid D)\right\vert.
\end{equation}

\emph{Proof.} By \hyperref[assumption:ind_generation]{Assumption 1}, $f(y_c) \perp f(y_r)$ given any $x$, with means $p_c(x) \coloneqq \mathbb{E}_{y\sim\pi_{\theta_c}(\cdot|x)}[f(y)]$ and $p_r(x)$ analogously. Conditioning on $x \in D$ and applying the law of total probability to each joint term in Equation~\ref{eq:acc_f} gives, e.g., $\mathbb{P}(f(y_c){=}1,f(y_r){=}0\mid D) = \mathbb{E}_{x\in D}[p_c(x)(1-p_r(x))]$; combining all terms as in the proof of \hyperref[prop:fixed_pair_bias]{Proposition 1} and using linearity of expectation yields Equation~\ref{eq:corollary_domain}.

For a fixed pair, the propensity gap is constant not just on dataset-wide average but \emph{within every domain}: it does not depend on $x$ beyond which domain $x$ falls in, and does not average out as $|\mathcal{D}| \to \infty$. This matters because intuitively spurious features are often domain-specific — e.g., \texttt{\textbackslash boxed\{\}} or \texttt{\#} markers in math — so a small marginal gap $\Delta_f(\theta_c,\theta_r)$ can mask large, opposite-signed, domain-conditional gaps that cancel only in aggregate. Since prompt domain is itself an observable covariate rather than a spurious feature the policy needs to ignore, this marginal cancellation offers no protection: a model can condition on domain implicitly and still exploit the domain-local shortcut in each region of the prompt distribution. This directly explains why, empirically, chosen and rejected completions from a single pair can become trivially separable using only surface-level features; we confirm this in Section~\ref{sec:style_bias}, where fixed model pairs differing sharply in a domain-specific stylistic propensity yield a very strong classifier. The same conditioning argument extends to the following \hyperref[prop:prop_pooling_generators]{Proposition 2}: pooling attenuates the bias only insofar as generator identity varies \emph{within} each domain, not merely across the dataset as a whole.

\paragraph{Assumption 2 (Pool sampling with $\mathcal{F}_l$-driven selection).} \phantomsection \label{assumption:pool_sampling_quality_driven}Now let $(\Theta_c^{(i)}, \Theta_r^{(i)})$ be drawn per-example from a pool of generators $\mathbb{M}$, with the chosen/rejected role at example $i$ assigned by a (possibly stochastic) selection criterion $c$ that depends on a completion only through the features in $\mathcal{F}_l$ (e.g., an external judge scoring safety).

\paragraph{Proposition 2 (Pooling attenuates the bias).}
\phantomsection\label{prop:prop_pooling_generators}
\emph{Under \hyperref[assumption:pool_sampling_quality_driven]{Assumption 2}, the dataset-level accuracy of a classifier using $f \in \mathcal{F}_s$ is}
\begin{equation}
\label{eq:prop_pooling_generators}
    \overline{\mathrm{Acc}_f} = \tfrac12 + \tfrac12\,\left\vert\mathbb{E}_i\big[\Delta_f(\Theta_c^{(i)}, \Theta_r^{(i)})\big]\right\vert
\end{equation}
\emph{obtained by averaging Equation~\ref{eq:prop1} over the per-example draw of $(\Theta_c^{(i)}, \Theta_r^{(i)})$, where $\Delta_f(\theta_c,\theta_r) \coloneqq p_f(\theta_c) - p_f(\theta_r)$ is the propensity gap. If, in addition, $f$ is uncorrelated with $\mathcal{F}_l$ at the population level — so that $c$'s dependence on $\mathcal{F}_l$ induces no dependence on $f$, and $\mathbb{E}_i[p_f(\Theta_c^{(i)})] = \mathbb{E}_i[p_f(\Theta_r^{(i)})]$ — the aggregate propensity gap vanishes and we have that $\overline{\mathrm{Acc}_f} = \tfrac12$.}

The key contrast with \hyperref[prop:fixed_pair_bias]{Proposition 1} is not only in expectation but in \emph{variance}: for a fixed pair, $\Delta_f(\theta_c,\theta_r)$ is a constant, so every example reinforces the same shortcut. Under pool sampling it is a random variable across examples — the model exhibiting $f$ more often is not always the one used to produce the 'chosen' — so no single feature threshold tracks the label consistently across $\mathcal{D}$, even if it correlates locally for some pairs. This is consistent with the substantially higher reward margins and pairwise accuracy observed for fixed model pairs regardless of the model trained, as shown in Section~\ref{sec:train_dynamics}.

We note that the aggregate propensity gap $\mathbb{E}_i[\Delta_f(\Theta_c^{(i)}, \Theta_r^{(i)})]$ need not vanish in general: if a feature in $\mathcal{F}_l$ correlates with a spurious feature at the population level (e.g., stronger models happen to write longer answers on average), pooling will not drive $\overline{\mathrm{Acc}_f}$ to $\tfrac12$. This residual case is qualitatively different from \hyperref[prop:fixed_pair_bias]{Proposition 1}'s bias, however: it reflects a property of the model population rather than an artifact of the sampling procedure, and, unlike the deterministic gap in the proposition, which is fixed by the choice of exactly two generators, it does not obviously worsen as the pool shrinks to a pair, nor improve as it grows. We leave a full characterization of this population-level confound to future work.

\subsection{Experimental Setup}

\subsubsection{Training}

\paragraph{Data.} To study the limitations of generating preference pairs from only two models, we adapt the DPO dataset used for \olmoTwoSft \citep{olmo20252olmo2furious} \href{https://huggingface.co/datasets/allenai/olmo-2-0425-1b-preference-mix}{\olmoDataset}, which originally relied on completions from multiple models ranked by an LLM-as-a-judge. To enable a direct 1-to-1 comparison, we re-generate all completions using models spanning various parameter sizes and model families. Following the heuristic in \citet{delta_learning_hyp}, we assign the chosen ($y_c$) and rejected ($y_r$) completions based strictly on the parameter size of the generating models. 
\paragraph{DPO.} We run one thousand training steps with a batch size of 128, sweeping learning rates and $\beta$ hyperparameters. We select the final ($\beta$, learning rate) configurations based on average downstream benchmark performance of a separate unseen evaluation set. We use the length-normalized loss defined in Equation \ref{eq:dpo_loss_len_norm}, which has been shown to outperform the original formulation and alleviate length bias~\citep{lambert2025tulu3pushingfrontiers}. More details are provided in Appendix \ref{app:training_details}.

\subsubsection{Models and Evaluation}

We select the training models and evaluation benchmarks according to three criteria. First, we aim to study models for which the preference signal is informative: the trained models should be strong enough to learn from the generated preference pairs, but not so strong that performance on the downstream benchmarks is already saturated. Second, we seek diversity in the models used to construct preference pairs, in order to test whether the conclusions depend on a particular model family or parameter scale. Third, we want the evaluation suite to reflect the capabilities relevant to the original dataset while also covering a broader range of skills and not saturated.
\paragraph{Training Model Selection.} We therefore select models along two dimensions: \emph{capability} and \emph{diversity}. In terms of capability, we focus on models with performance broadly comparable to \olmoTwoSft, the model for which the dataset's prompt distribution was originally designed. In terms of diversity, we include models from different families and parameter scales. Thus, we consider the following Hugging Face checkpoints: \qwenTwoSmall~\citep{qwen2}, \gemmaTwoTiny~\citep{gemma2}, \aya~\citep{aya23}, and \mistral~\citep{mistral7b}, \qwenOneHalfSmall and \qwenOneHalfBig~\citep{qwen1.5}.
\paragraph{Generator Model Selection.} We use a staged approach to selecting generator models. First, we conduct preliminary experiments with \qwenThreeSmall and \qwenThreeTiny, which provide a cost-effective setting while maintaining a meaningful parameter gap, following the 2-3B parameter differences used in \citet{delta_learning_hyp}. We then test the robustness of our findings across model families and parameter scales, using model size as the heuristic for constructing preference pairs always within the same family where larger parameter count reliably predicts benchmark performance within that lineage as done in the reference work. Finally, we expand the generator pool beyond two models to investigate whether sampling completions from multiple models can improve performance under the same size-based preference heuristic. The specific models used in each experiment are described in the corresponding results sections, with the complete list provided in Appendix~\ref{sec:models_generator}.
\paragraph{Benchmark selection.} We evaluate models on TruthfulQA~\citep{tqa}, ARC-Challenge~\citep{arc_challenge}, IFEval~\citep{ifeval}, Math-500~\citep{math_500}, GSM8K~\citep{gsm8k}, Human-Eval~\citep{human_eval}, and MBPP~\citep{mbpp}. To select the best hyperparameters we use the training set of GSM8K, the validation set of ARC-Challenge and IFBench~\citep{IFBench} since IFEval lacks separate split. Detailed evaluation settings are provided in Appendix Table~\ref{tab:benchmarks}; Appendix Table~\ref{tab:model_evals_qwen4b_qwen0.6b} confirms that performance does not saturate on these benchmarks across the model scales considered.

\section{Results}

In the following sections, we show comprehensively how using just a pair of models to generate DPO data leads to a significantly lower downstream performance and training instability. 

\subsection{Delta Learning doesn't match LLM-as-a-judge}
\label{sec:delta_learning_worse_than_llm_judge}

\begin{figure}[t]
  \centering
  \begin{subfigure}[t]{0.48\columnwidth}
    \centering
    \includegraphics[width=\textwidth]{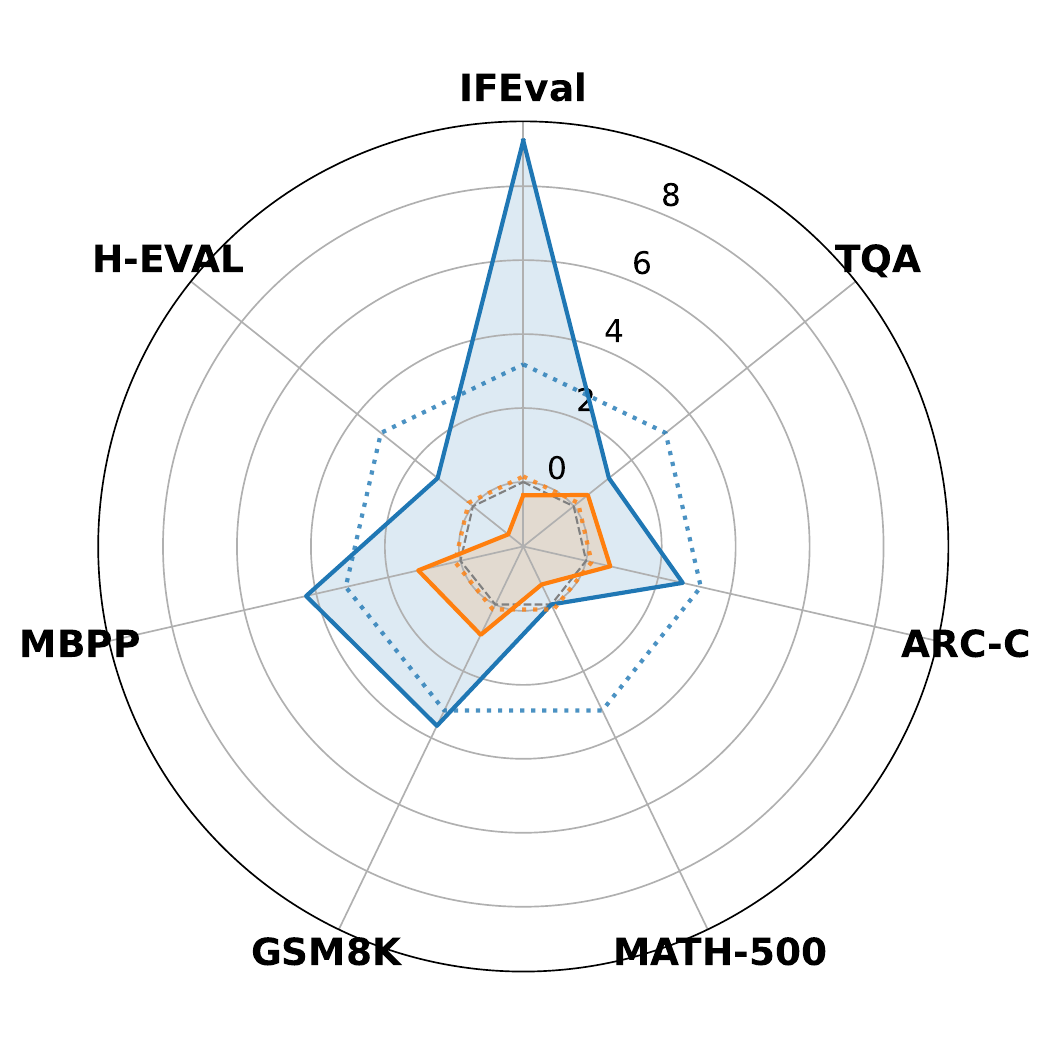}
    \caption{\qwenTwoSmall}
    \label{subfig:qwen2_15b_qwen3_4B-0.6B}
  \end{subfigure}
  \hfill
  \begin{subfigure}[t]{0.48\columnwidth}
    \centering
    \includegraphics[width=\textwidth]{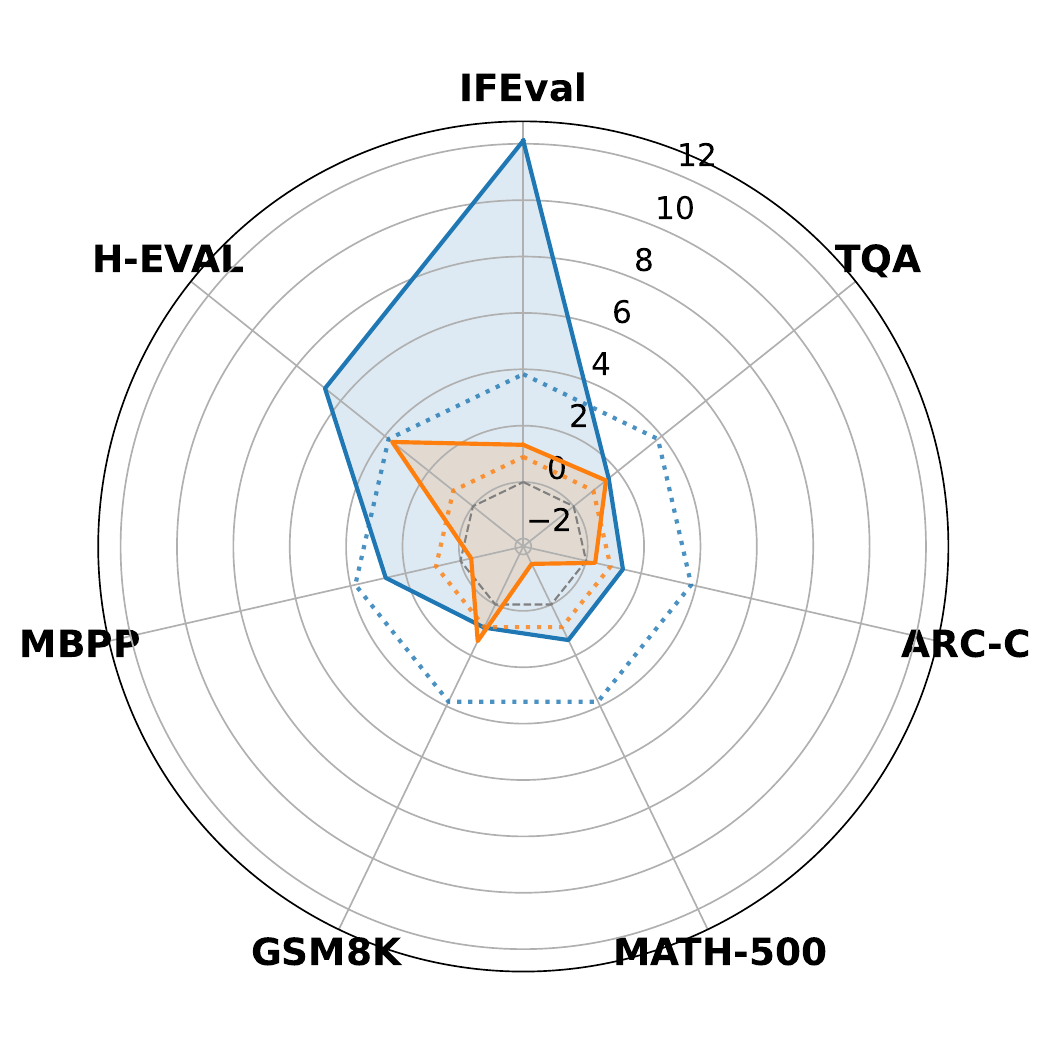}
    \caption{\gemmaTwoTiny}
    \label{subfig:gemma2_2b_qwen3_4B-0.6B}
  \end{subfigure}

  \vspace{0.4em} %

  \begin{subfigure}[t]{0.48\columnwidth}
    \centering
    \includegraphics[width=\textwidth]{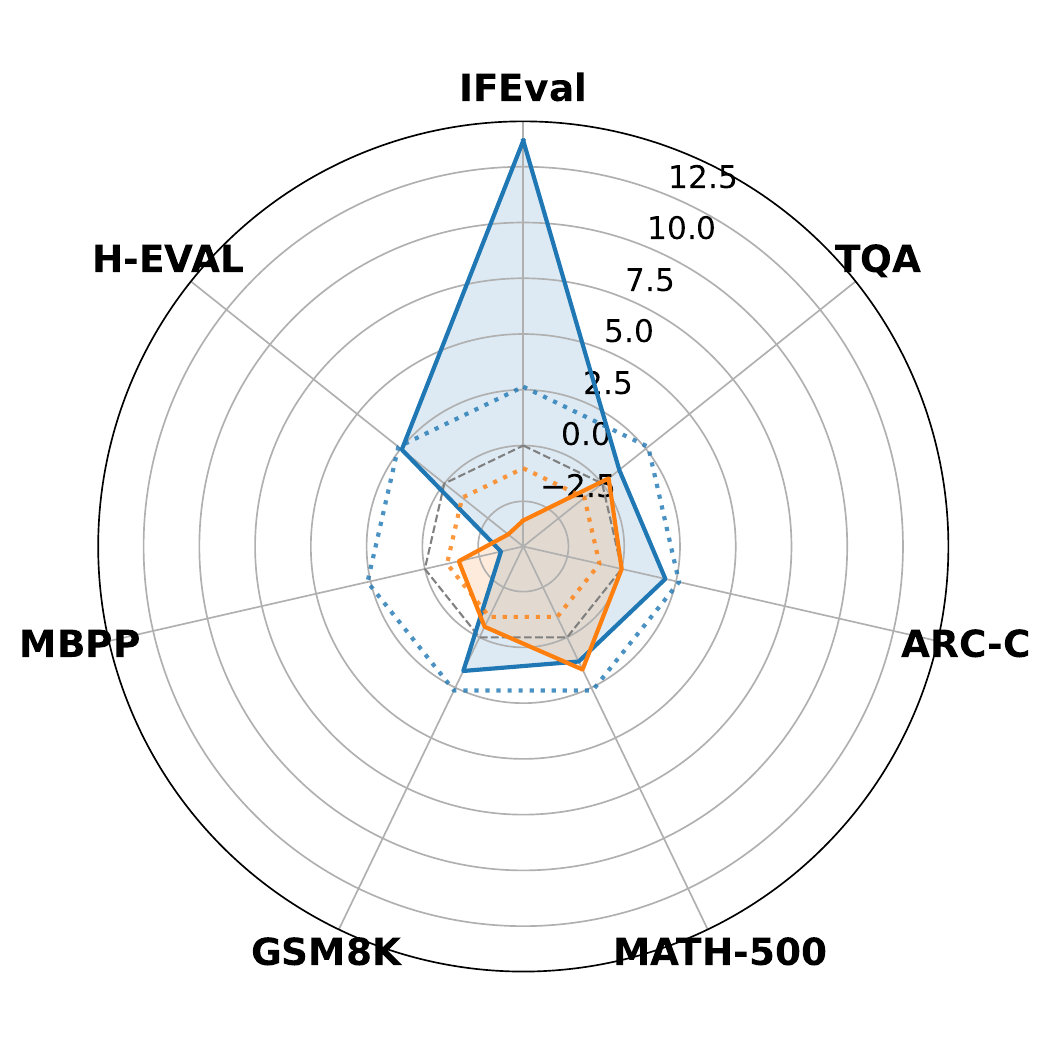}
    \caption{\qwenOneHalfSmall}
    \label{subfig:qwen15_4b}
  \end{subfigure}
  \hfill
  \begin{subfigure}[t]{0.48\columnwidth}
    \centering
    \includegraphics[width=\textwidth]{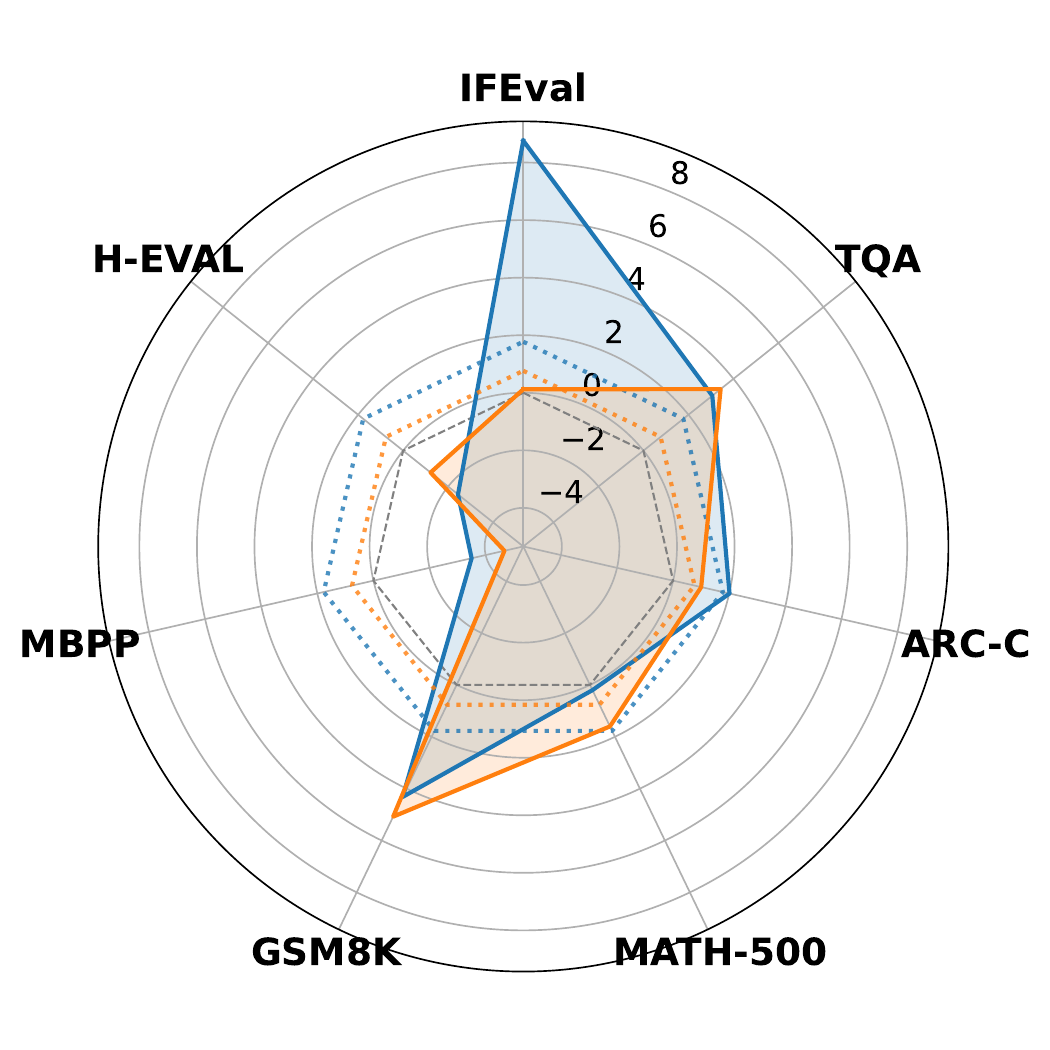}
    \caption{\aya}
    \label{subfig:aya23_8b}
  \end{subfigure}%

  \vspace{0.4em}

  \begin{subfigure}[t]{0.48\columnwidth}
    \centering
    \includegraphics[width=\textwidth]{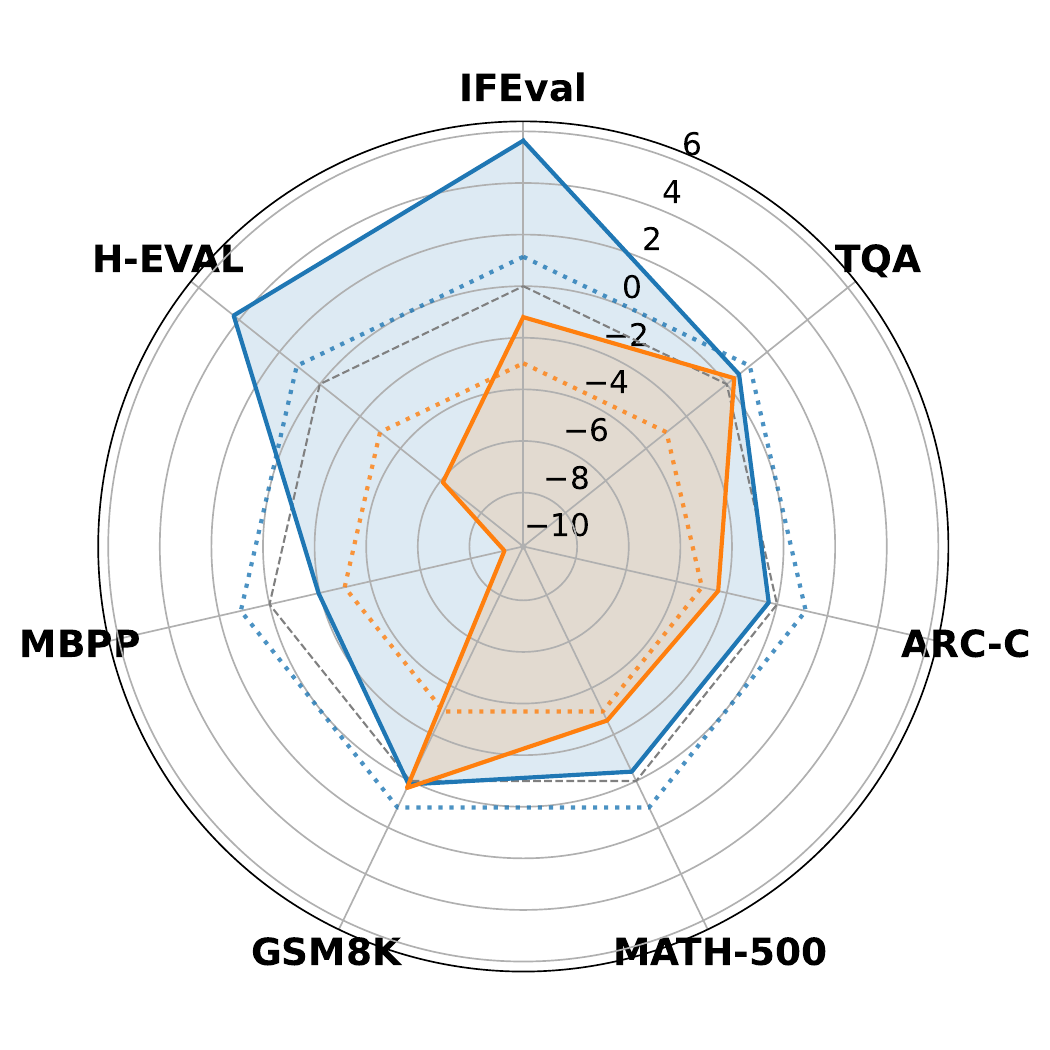}
    \caption{\qwenOneHalfBig}
    \label{subfig:qwen15_7b}
  \end{subfigure}%
  \hfill
  \begin{subfigure}[t]{0.48\columnwidth}
    \centering
    \includegraphics[width=\textwidth]{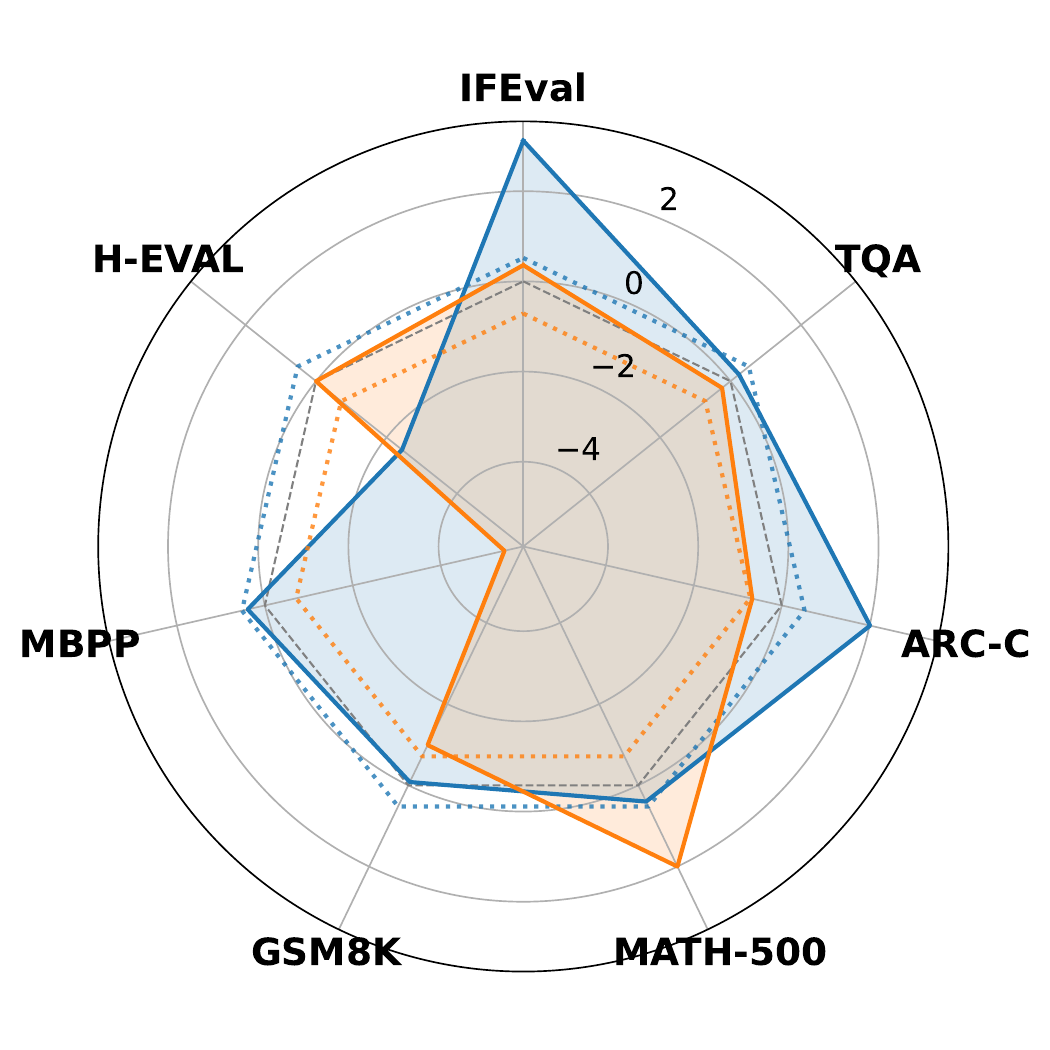}
    \caption{\mistralPlot}
    \label{subfig:mistral_7b}
  \end{subfigure}%

  \caption{\textbf{Delta learning underperforms multiple-models.} Relative score changes across benchmarks and models using \olmoDataset as base dataset. Performance is compared between data generated using \textcolor{tabblue}{multiple models (blue)} versus \textcolor{taborange}{a single pair (orange)} (\qwenThreeSmall vs \qwenThreeTiny), with dotted lines indicating average performance.}
  \label{fig:main_radar_plots}
\end{figure}

First, we aim to investigate whether the performance of models of different families and sizes change when training on a fixed prompt distribution while using different generators. Figure \ref{fig:main_radar_plots} shows how regardless of the family and size, all models on average and on most benchmarks improve the performance, while when using \qwenThreeSmall and \qwenThreeTiny for half the performance degrades and for the other half it remains unaltered or lags behind the original dataset by a large margin. 

\begin{figure}[t]
  \centering
  
  \begin{subfigure}[t]{0.48\columnwidth}
    \centering
    \includegraphics[width=\textwidth]{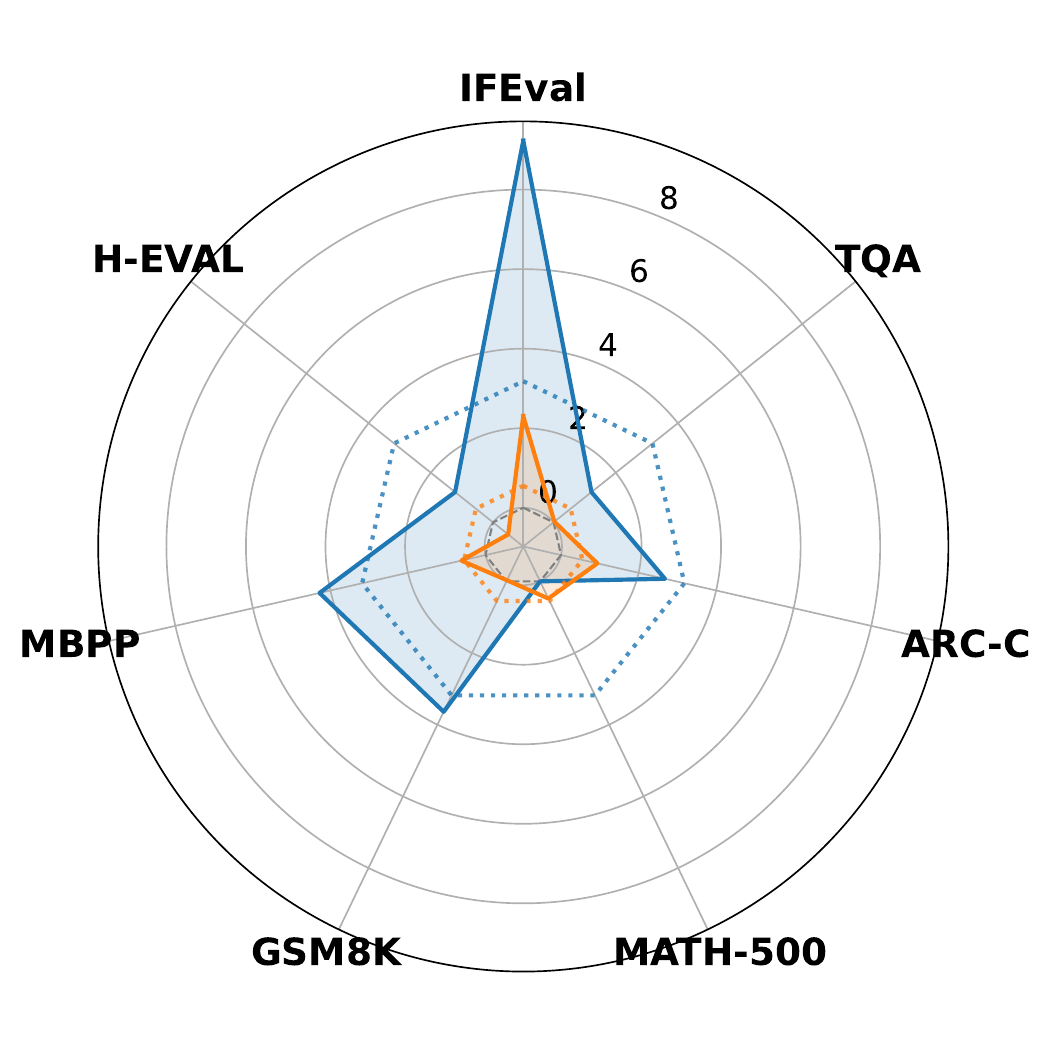}
    \caption{\qwenTwoSmall}
    \label{subfig:qwen2_15b_pairs_ablation}
  \end{subfigure}
  \hfill
  \begin{subfigure}[t]{0.48\columnwidth}
    \centering
    \includegraphics[width=\textwidth]{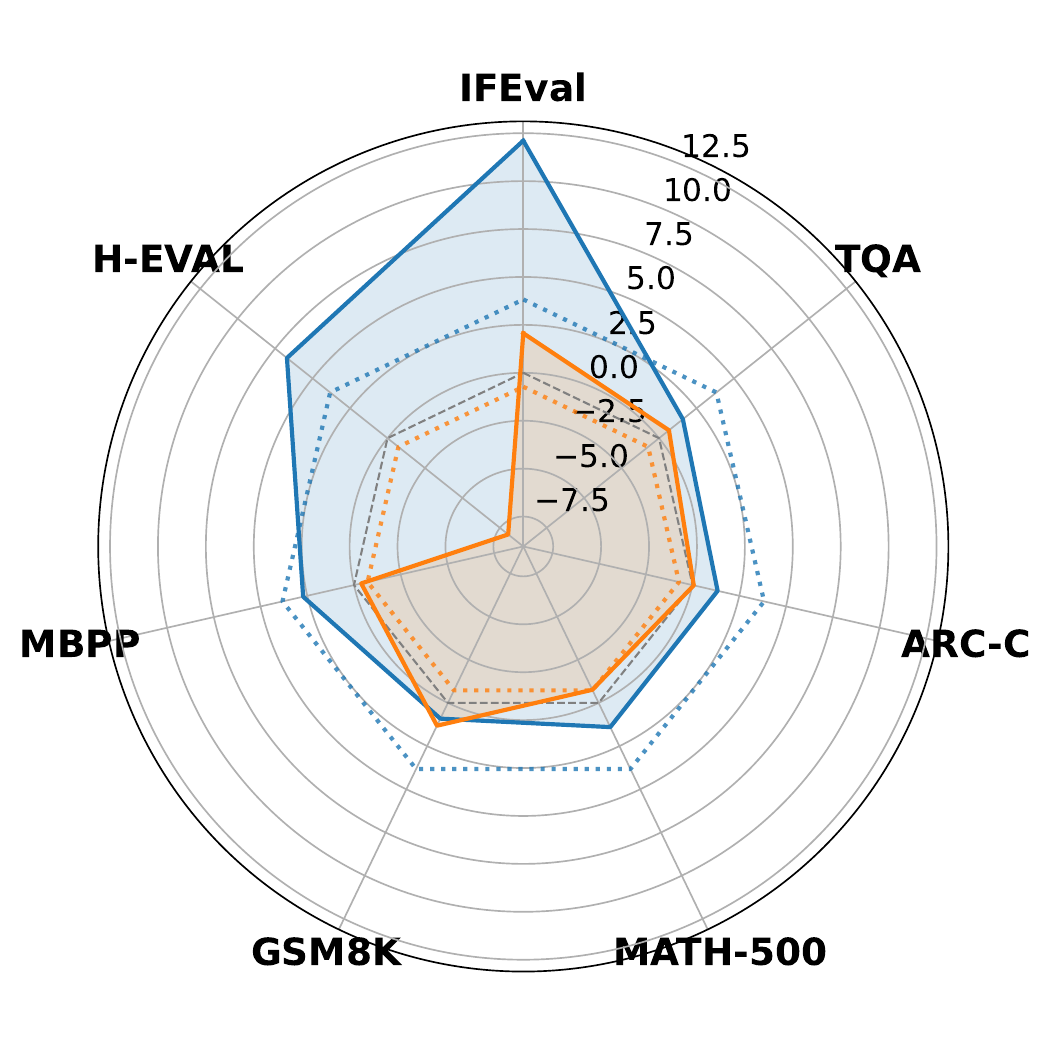}
    \caption{\gemmaTwoTiny}
    \label{subfig:gemma2_2b_pairs_ablation}
  \end{subfigure}

  \vspace{0.4em} %

  \begin{subfigure}[t]{0.48\columnwidth}
    \centering
    \includegraphics[width=\textwidth]{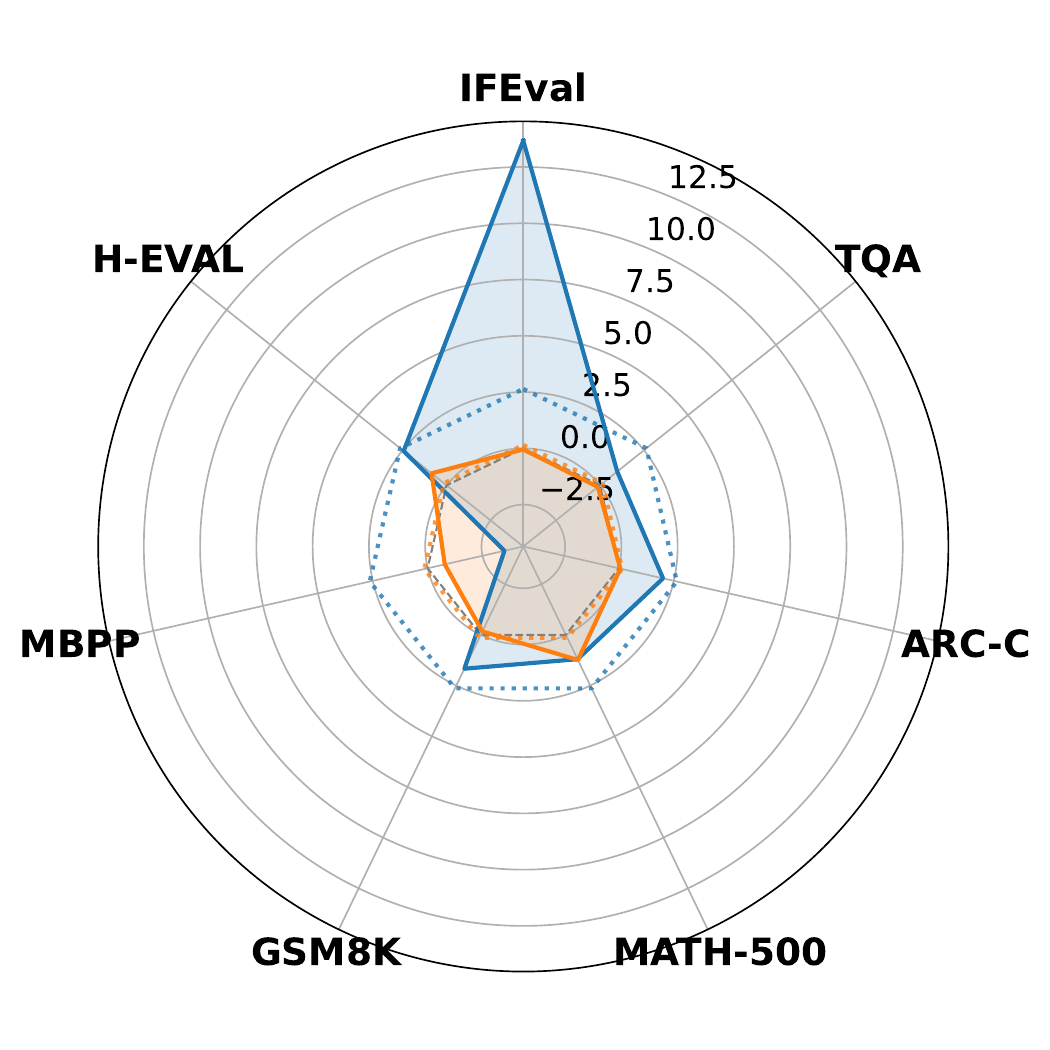}
    \caption{\qwenOneHalfSmall}
    \label{subfig:qwen15_4b_pairs_ablation}
  \end{subfigure}
  \hfill
  \begin{subfigure}[t]{0.48\columnwidth}
    \centering
    \includegraphics[width=\textwidth]{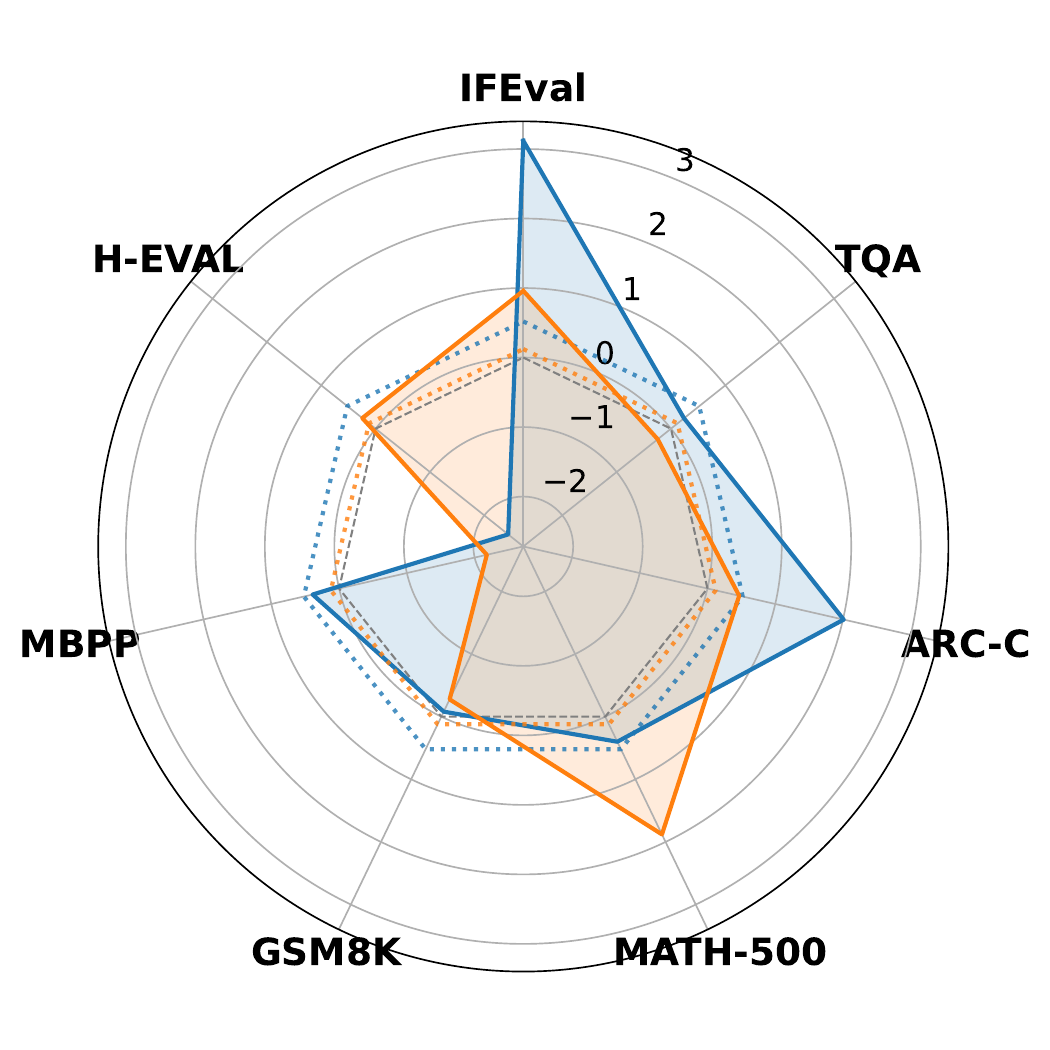}
    \caption{\mistralPlot}
    \label{subfig:mistral_7b_pairs_ablation}
  \end{subfigure}%

  \caption{\textbf{Lower performance doesn't depend on the generators.} Relative score changes across benchmarks. Performance is compared between data generated using \textcolor{tabblue}{multiple models} versus \textcolor{taborange}{a single pair}, with dotted lines indicating average performance. \textcolor{taborange}{Thick orange line} represents the average per-benchmark across pairs used.}
  \label{fig:pair_ablations_radar_plots}
\end{figure}

To verify that the low performance is not merely an artifact of the specific pair chosen, we run additional experiments using a subset of the previous trained models, while using the following models as pairs: (\qwenThreeBig, \qwenThreeTiny), (\olmoTwoBig, \olmoTwoSmall), (\olmoTwoSmall, \olmoTwoTiny) and (\gemmaTwoBig, \gemmaTwoSmall). Figure \ref{fig:pair_ablations_radar_plots} shows how, while we can observe a small improvement due to the variance between the pairs used, on average the score remains much lower compared to when using the original dataset thus proving that there's an intrinsic problem in the completions. Tables \ref{tab:model_evals_qwen4b_qwen0.6b} and \ref{tab:pairs_ablations} in the appendix show the exact evaluation numbers.

\subsection{Does data-model alignment matter?}

Before addressing the problem behind the inferior improvement, we seek to remove an additional spurious variable that might distort the analysis. Specifically, we aim to verify that the nature of the results doesn't simply stem from the fact that the prompt distribution wasn't originally designed for the models tested, which might make training harder especially when training with an off-policy algorithm such as DPO~\citep{is_dpo_superior_to_ppo, what_matters_data_dpo}. Thus, we run DPO with \olmoTwoSft on both the original dataset and on the version with completions generated with all the models used in the previous section and evaluate on the subset of benchmarks used in the original paper~\citep{olmo20252olmo2furious}. Table \ref{tab:olmo2_sft_dpo_ablation_summary} shows how the data distribution doesn't appear to be an important variable, as training on the original dataset yields strong improvements on all benchmarks, while it's clearly very unstable when using only a pair of models and average degrades. Results for each pair of models generator are shown in the appendix in Table \ref{tab:olmo2_sft_dpo_ablation}. 

\begin{table}[ht]
\centering
\caption{\textbf{Delta pair doesn't underperform because of model-dataset alignment.} Benchmark deltas using the original \olmoDataset dataset and different model pairs to generate the completions.}
\label{tab:olmo2_sft_dpo_ablation_summary}
\resizebox{\columnwidth}{!}{%
\setlength{\tabcolsep}{3pt}
\begin{tabular}{l c c c c c}
\toprule
\textbf{Model / Dataset} & \textbf{IFEval} & \textbf{TQA} & \textbf{MATH-500} & \textbf{GSM8K} & \textbf{Avg} \\
\midrule
\olmoTwoSft & 61.63\% & 26.93\% & 14.60\% & 51.86\% & 38.75\% \\
\olmoDataset & \pos{7.79\%} & \pos{4.53\%} & \pos{0.60\%} & \pos{6.29\%} & \textbf{\pos{4.80\%}} \\
\avgPairs & \negval{-0.31\%} & \pos{0.34\%} & \negval{-1.12\%} & \pos{0.50\%} & \textbf{\negval{-0.14\%}} \\
\bottomrule
\end{tabular}%
}
\end{table}

\subsection{Spurious Features Analysis}
\label{sec:spurious_analysis}

Since neither the data distribution nor the choice of model generators appears to be entirely responsible for the low performance of DPO, we next conduct an analysis of the generated data and the training dynamics. In Section \ref{sec:delta_learning_limits}, we theoretically showed that spurious features are more likely to emerge when data are generated from a small number of model pairs rather than from a larger and more diverse pool of generators. In this section, we empirically investigate this phenomenon and show that, for some tasks, such biases can become particularly pronounced.

\subsubsection{Training Dynamics}
\label{sec:train_dynamics}

Based on \citet{delta_learning_hyp} for a fixed prompt distribution, data generated from a single pair of models should provide a preference signal comparable to that obtained from a pool of models and an external judge. In particular, if the chosen response is truly preferred because of its underlying task quality, then both procedures should expose the model to essentially the same signal. To formalize this, let the reward with a fixed prompt be

\begin{equation}
\label{eq:reward}
    r_\theta(y) = \frac{\beta}{|y|}\log\frac{\pi_\theta(y\mid x)}{\pi_{\mathrm{ref}}(y\mid x)}
\end{equation}

The reward margin is then $r_\theta(y_w)-r_\theta(y_l)$, while reward accuracy is $\frac{1}{N}\sum_{i=1}^N \mathbbm{1}[r_\theta(y_w^i)>r_\theta(y_l^i)]$. We use \emph{reward hacking} to describe the policy increasing its reward margin and accuracy during training without a corresponding improvement in downstream benchmark performance. Since the prompt distribution and the underlying quality difference between chosen and rejected responses are held fixed across the two settings, differences in reward metrics can reveal whether the policy is learning the intended preference or exploiting non-causal, generator-dependent signals.

\begin{figure}[t]
  \centering
  \begin{subfigure}[t]{0.48\columnwidth}
    \centering
    \includegraphics[width=\textwidth]{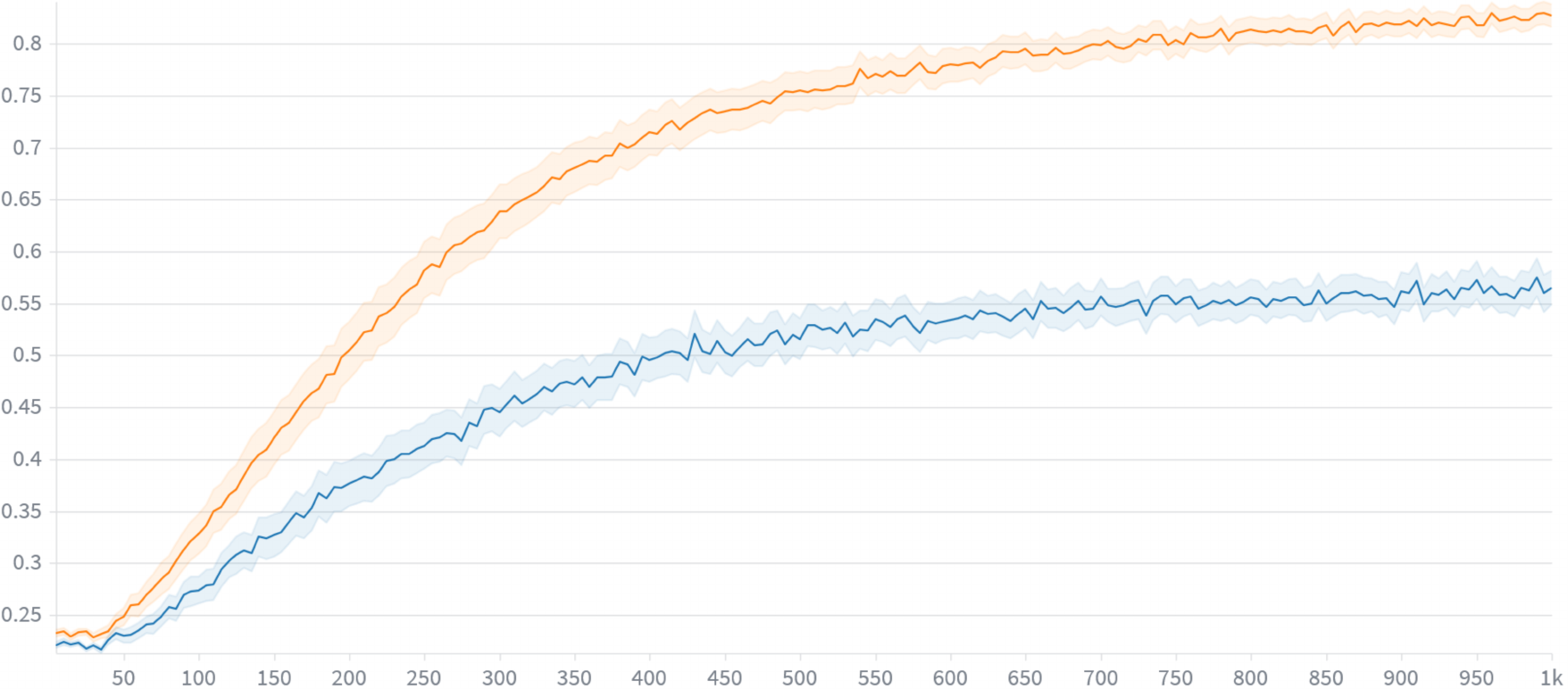}
    \caption{Rewards accuracy}
    \label{subfig:rewards_accuracy}
  \end{subfigure}
  \hfill
  \begin{subfigure}[t]{0.48\columnwidth}
    \centering
    \includegraphics[width=\textwidth]{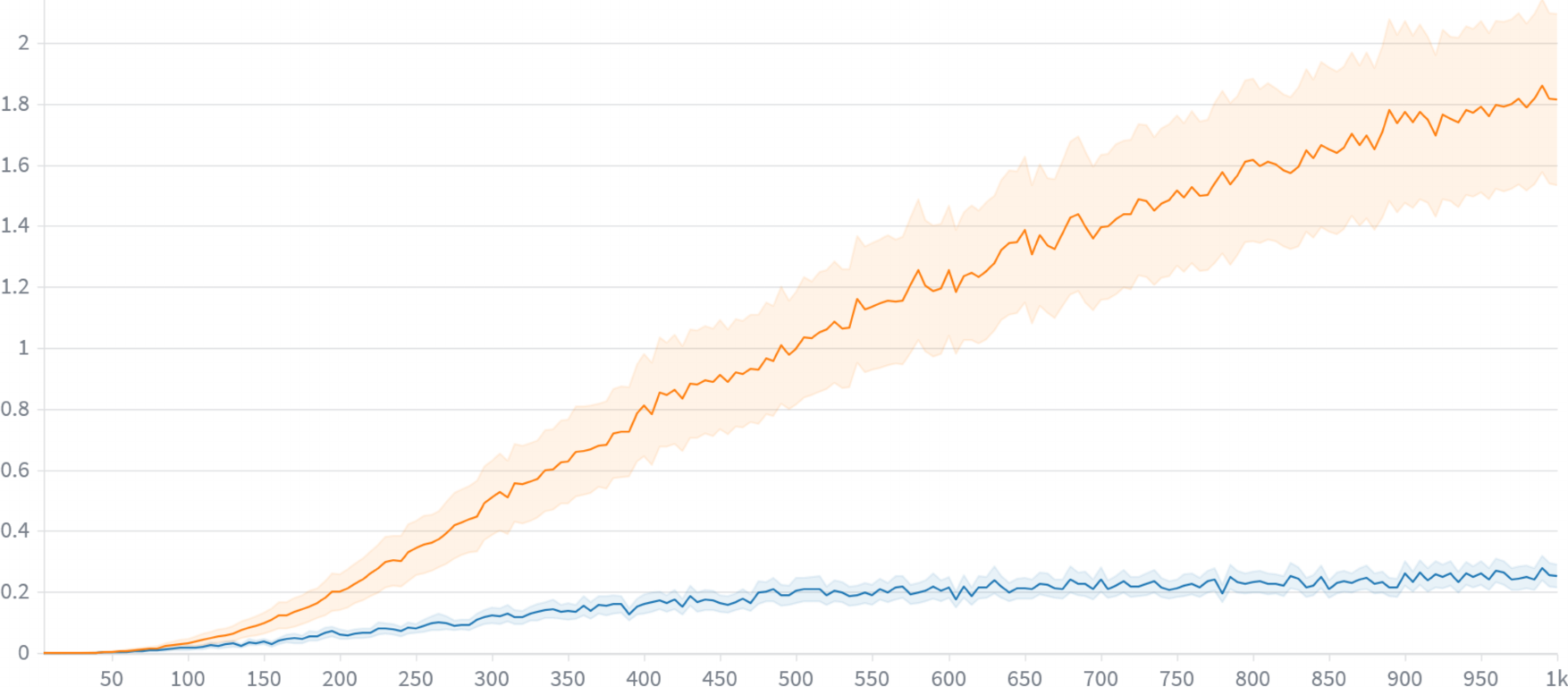}
    \caption{Rewards margins}
    \label{subfig:rewards_margins}
  \end{subfigure}

  \caption{\textbf{Training statistics.} Mean reward accuracy and margin $\pm$ standard error across models and hyperparameters for \textcolor{tabblue}{\olmoDataset} and \textcolor{taborange}{one pair}.}
  \label{fig:train_stats}
\end{figure}

If the signal learned from the single-pair data is primarily attributable to the intended preference between $y_w$ and $y_l$, we would therefore expect its training dynamics to be comparable to those obtained from the model pool and judge. Figure~\ref{fig:train_stats} shows the opposite: it reports averages across the models, generator pairs, and learning rates considered in Section~\ref{sec:delta_learning_worse_than_llm_judge}, with reward accuracy differing by approximately $27\%$ between the two settings. This substantial gap suggests that the single-pair construction exposes additional systematic signals beyond the intended preference, causing reward hacking and hindering training improvement.

\subsubsection{Length Bias}

Length difference is the most prominent and easily quantifiable completion bias. As shown in Table~\ref{tab:length_distribution}, length disparity provides a trivial shortcut for preference learning in the Qwen and Gemma model pairs. However, this pattern does not hold for the Olmo2 pairs, necessitating the study of more nuanced biases.

\begin{table}[ht]
\centering
\caption{\textbf{Length Bias.} Length completion characters difference between pairs used in Section \ref{sec:delta_learning_worse_than_llm_judge}.}
\label{tab:length_distribution}
\resizebox{\columnwidth}{!}{%
\setlength{\tabcolsep}{3pt}
\begin{tabular}{l c c c}
\toprule
\textbf{Chosen/Dataset} & \textbf{Rejected} & \textbf{Pct Longer\%} & \textbf{Median $\Delta$} \\
\midrule
\olmoDataset & - & 55.25\% & 64 \\
\midrule
\qwenThreeSmall & \qwenThreeTiny & 81.66\% & 510 \\
\qwenThreeBig & \qwenThreeTiny & 81.49\% & 524 \\
\gemmaTwoBig & \gemmaTwoSmall & 86.96\% & 723 \\
\olmoTwoSmall & \olmoTwoTiny & 47.72\% & 0.0 \\
\olmoTwoBig & \olmoTwoTiny & 43.84\% & -8.0 \\
\bottomrule
\end{tabular}%
}
\end{table}

\subsubsection{Stylistic Bias}
\label{sec:style_bias}

We focus on data from one of the Olmo2 pair with minimal length bias that still produces sub-par downstream performance, and another pair used to highlight how extreme and problematic this bias can become. 

We evaluate format bias on GSM8K by extracting stylistic features, such as bolding and hash delimiters,  while excluding length, which was analysed previously and controlled via a length-normalized loss that alleviates this issue. One might worry that some markers belong to the expected answer format rather than being spurious. This does not undermine the analysis: if a marker is genuinely required by the task (e.g. "Write the answer after \#\#\#"), a well-calibrated signal should push the policy to always produce it, so it would appear in both chosen and rejected completions and carry no discriminative power. Its strong separability here instead indicates inconsistent use across completions which is a tracker for generator identity rather than task compliance. 

\begin{figure}[t]
  \centering
  
  \begin{subfigure}[t]{0.48\columnwidth}
    \centering
    \includegraphics[width=\textwidth]{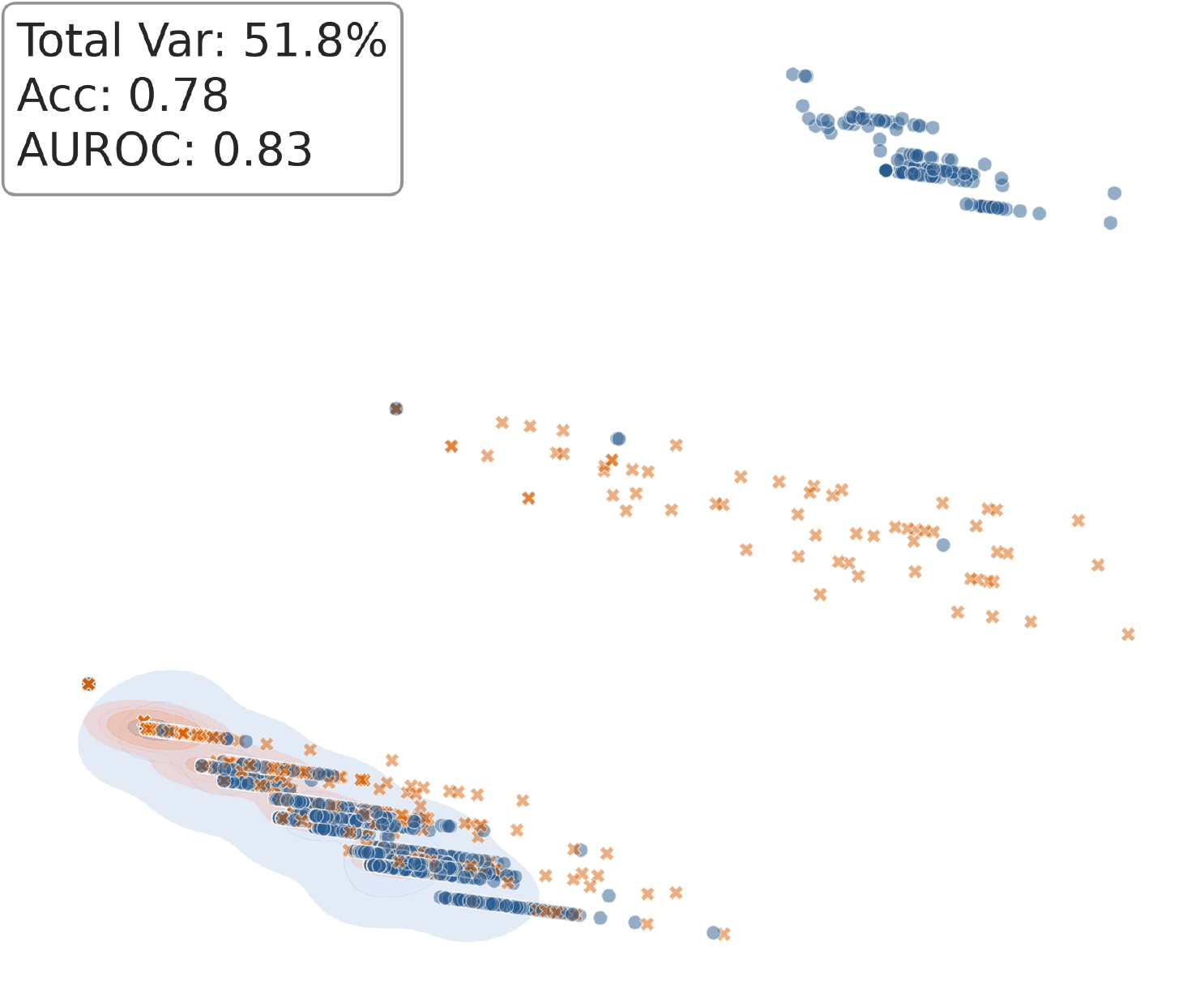}
    \caption{\centering \textcolor{chosenColor}{Olmo2 7B}-\textcolor{rejectedColor}{Olmo2 1B}}
    \label{subfig:olmo2_7b_1b_pca}
  \end{subfigure}
  \hfill
  \begin{subfigure}[t]{0.48\columnwidth}
    \centering
    \includegraphics[width=\textwidth]{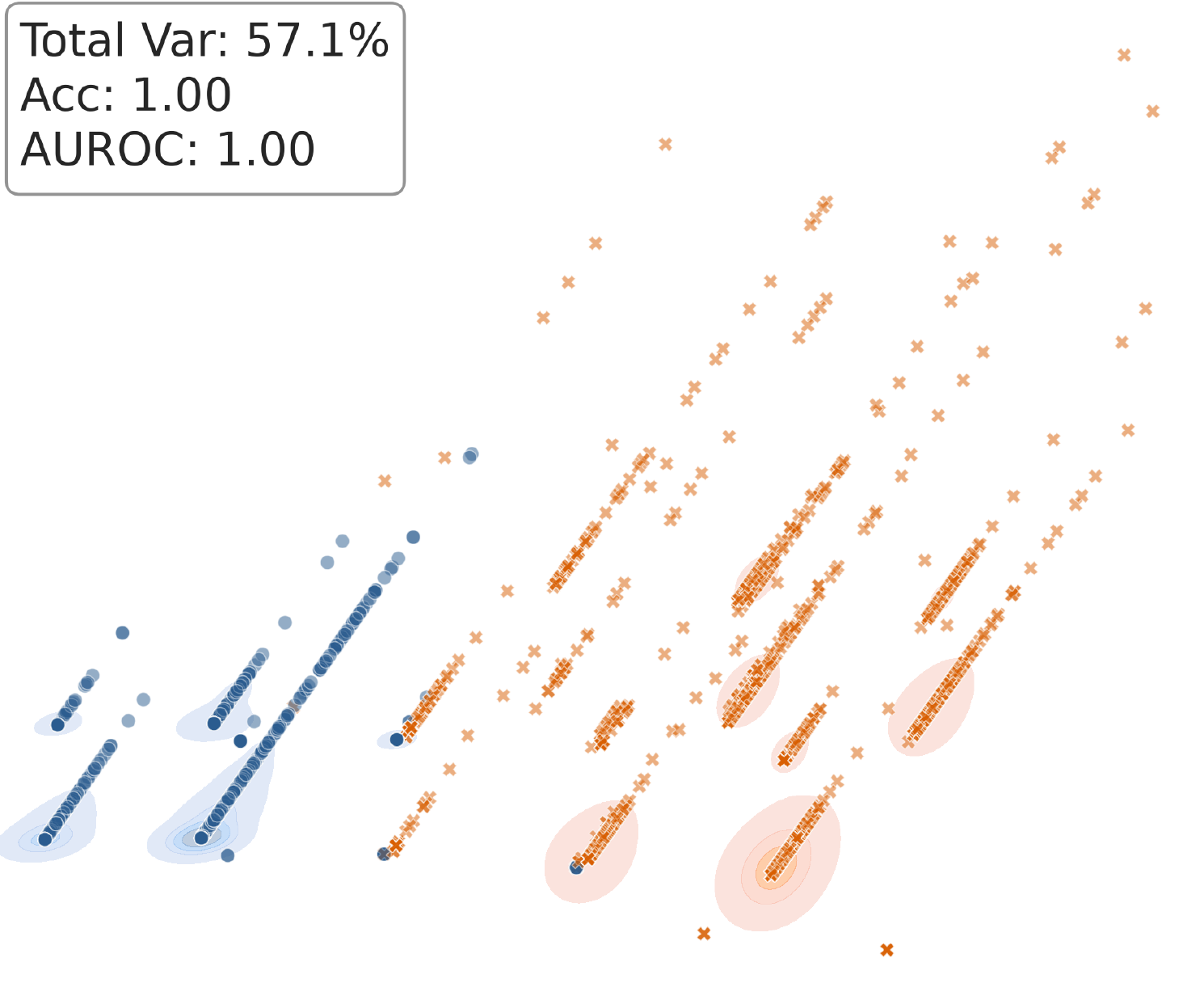}
    \caption{\centering \textcolor{chosenColor}{Gemma2 27B}-\textcolor{rejectedColor}{Olmo2 1B}}
    \label{subfig:gemma27_olmo_1b_pca}
  \end{subfigure}

  \vspace{0.4em}

  \caption{\textbf{Style Bias.} PCA of model pairs showing trivial separability with Logistic Regression scores and Explained Variance in the top-left.}
  \label{fig:style_pca}
\end{figure}

To test if stylistic shortcuts drive reward hacking, we map completions to normalized style vectors and project them into 2D via PCA. As shown in Figure~\ref{fig:style_pca}, the features exhibit clear separability, with the first two principal components explaining $51.8\%$ and $57.1\%$ of the total variance in the first and second datasets, respectively. Moreover, a logistic regression classifier on these features achieves 5-fold CV AUROC of $83\%$ and $100\%$. This indicates that stylistic features account for most of the structural variance between model outputs in this setting, consistent with their acting as a shortcut that drives the downstream instability and performance degradation observed in DPO. See Appendix~\ref{app:style_bias_empirical} for the full list of style features considered and additional details.

Directly intervening on these features during training would establish a causal link with downstream performance, but doing so is impractical on a large, diverse dataset like \olmoDataset, where the relevant spurious features vary across domains and cannot be enumerated or controlled in advance. We therefore test the mechanism indirectly: if these shortcuts stem from generator-dependent biases, increasing generator diversity without an external judge should weaken them and improve performance.

\subsection{Increasing Generators Diversity}
\label{sec:improve_delta_learning}

Having shown that a single generator pair is suboptimal, we now ask the complementary question: is increasing generator diversity alone, without an external judge, enough to mitigate the problem? \hyperref[prop:prop_pooling_generators]{Proposition 2} predicts that drawing chosen and rejected completions from a larger pool of generators should weaken the shortcut of \hyperref[prop:fixed_pair_bias]{Proposition 1}. We therefore primarily compare against the single-pair baseline rather than against the original \olmoDataset, which also differs in its ranking criterion. Our goal is not to close the gap with judge-supervised data, but to test whether diversity by itself yields a consistent improvement over single pairs. Note however that this approach is incremental: multi-model generation is a prerequisite for judge-based ranking, allowing practitioners to incorporate an external judge if needed.

\paragraph{Dataset Generation with Multiple Models.} Moving beyond a single model family breaks the size-based heuristic, since parameter count no longer reliably predicts completion quality across families (e.g., \gemmaTwoTiny outperforms \qwenOneHalfSmall despite having half the parameters; Table~\ref{tab:model_evals_qwen4b_qwen0.6b}). Using a judge would confound generators diversity with ranking criterion, potentially altering the choice even within a family. Therefore we assume access to benchmarks that proxy models' relative performance on the training domains. This holds for \olmoDataset, which is mainly composed of Instruction Following, Math and Coding data, so we rank models by their performance on IFEval, GSM8K and MBPP. We divide them into three tiers (high, mid, low) and, for each example, sample one of the pairings (high, mid), (high, low) or (mid, low), then draw one model uniformly from each tier. The result is a dataset whose completions come from nine different models. We note that it is the diversity of generators, not the benchmark-guided ranking, that \hyperref[prop:prop_pooling_generators]{Proposition 2} predicts will attenuate the bias; the ranking only replaces the size heuristic as a source of chosen/rejected ordering.

Since the ranking benchmarks also belong to our evaluation suite, we clarify that this is not data leakage, which would require test-set information to reach the training data through sample overlap or sample-level statistics. Neither occurs: the prompts in \olmoDataset{} contain no sample from these benchmarks, and the benchmarks are used only to rank the \emph{generators} by aggregate domain-level performance, a property of the models rather than of any test sample. The coarse ranking decides which model writes each completion, but carries no information about the benchmark samples themselves.

\begin{figure}[htbp]
  \centering
  
  \begin{subfigure}[b]{0.48\columnwidth}
    \centering
    \includegraphics[width=\textwidth]{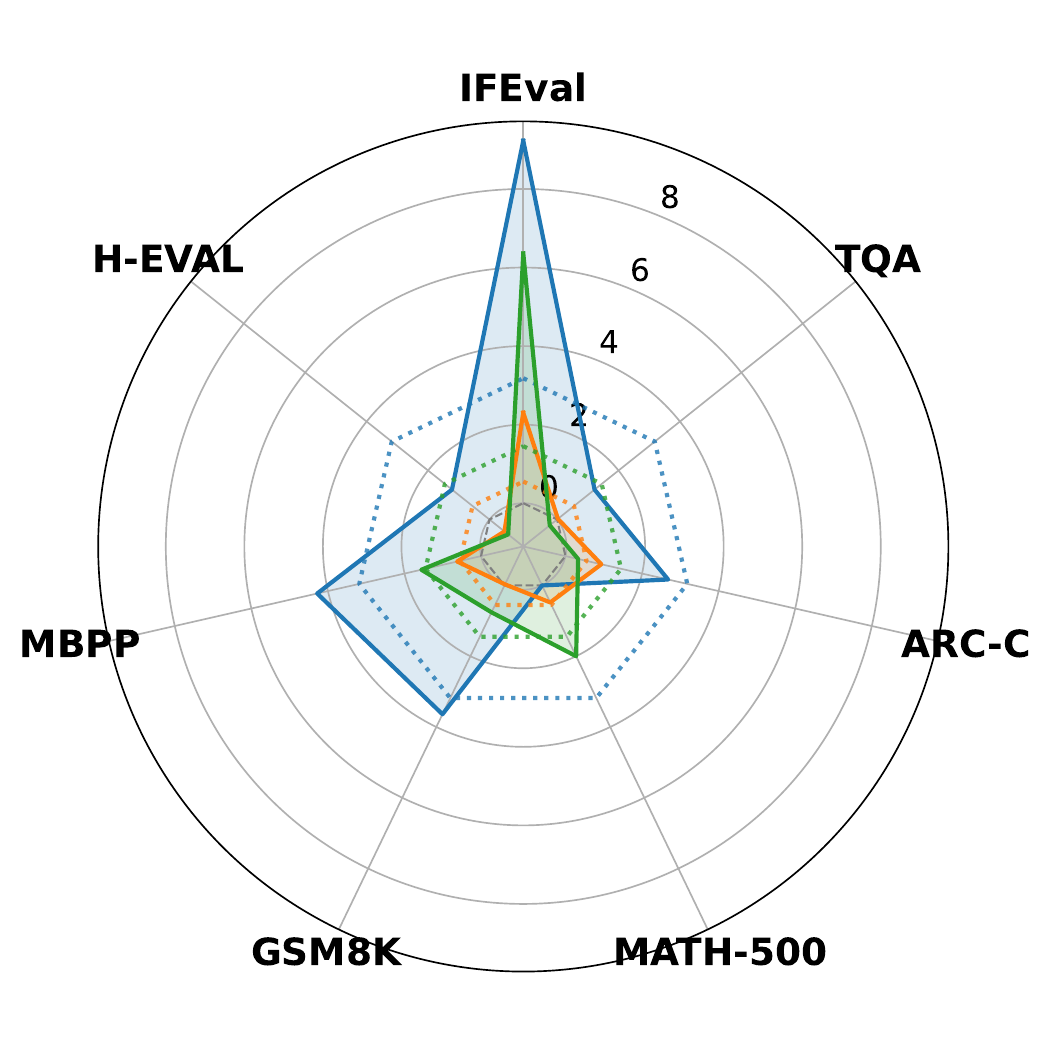}
    \caption{\qwenTwoSmall}
    \label{subfig:qwen2_15b_more_models}
  \end{subfigure}
  \hfill
  \begin{subfigure}[b]{0.48\columnwidth}
    \centering
    \includegraphics[width=\textwidth]{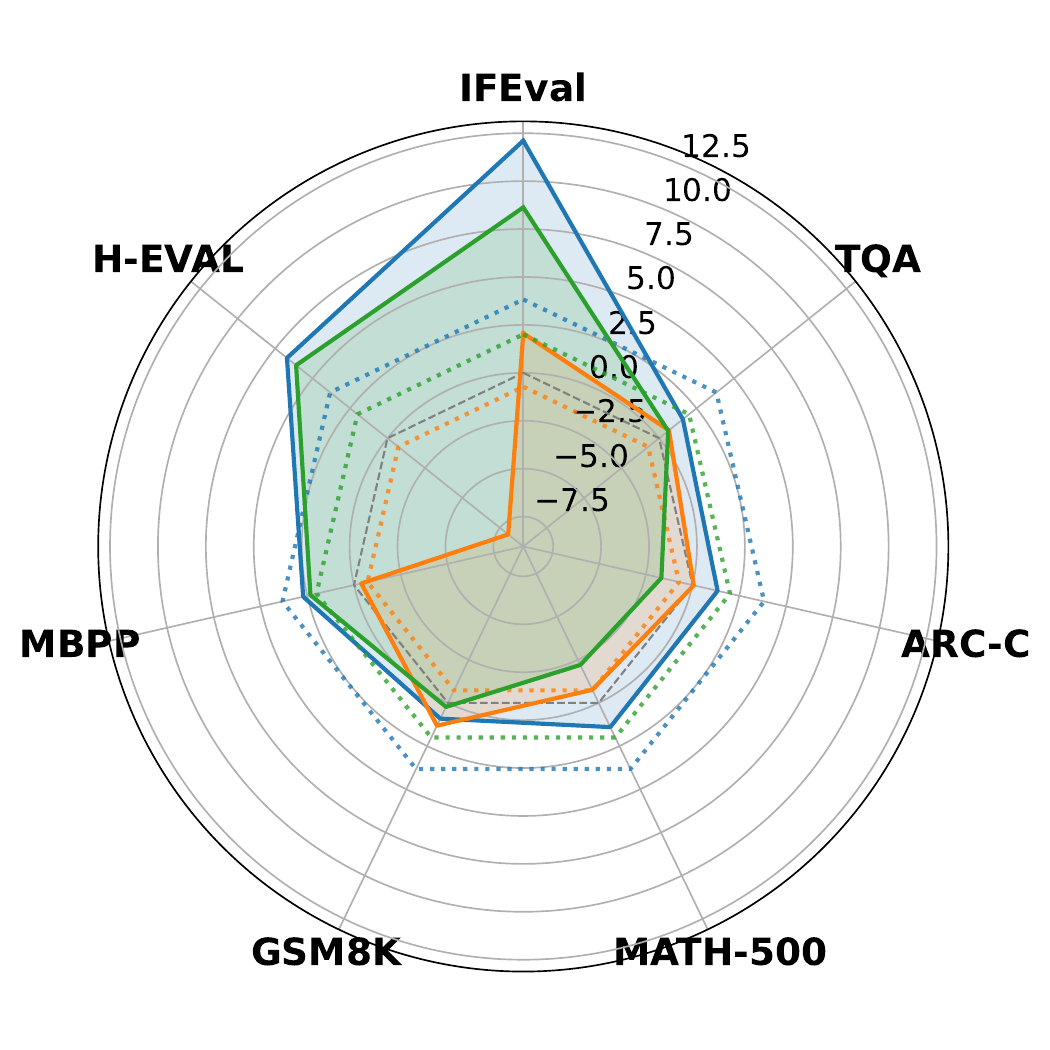}
    \caption{\gemmaTwoTiny}
    \label{subfig:gemma2_2b_more_models}
  \end{subfigure}

  \vspace{0.4em} %

  \begin{subfigure}[b]{0.48\columnwidth}
    \centering
    \includegraphics[width=\textwidth]{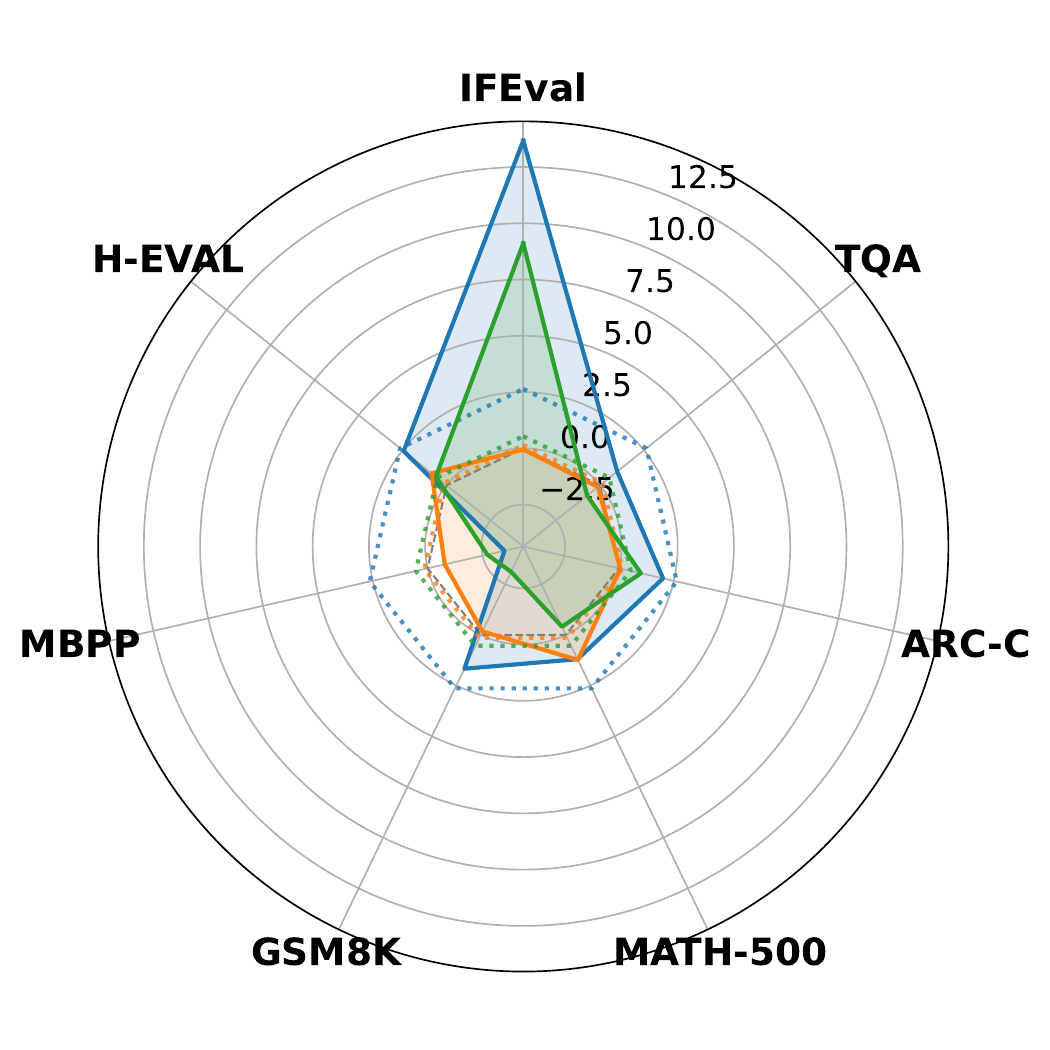}
    \caption{\qwenOneHalfSmall}
    \label{subfig:qwen15_4b_more_models}
  \end{subfigure}
  \hfill
  \begin{subfigure}[b]{0.48\columnwidth}
    \centering
    \includegraphics[width=\textwidth]{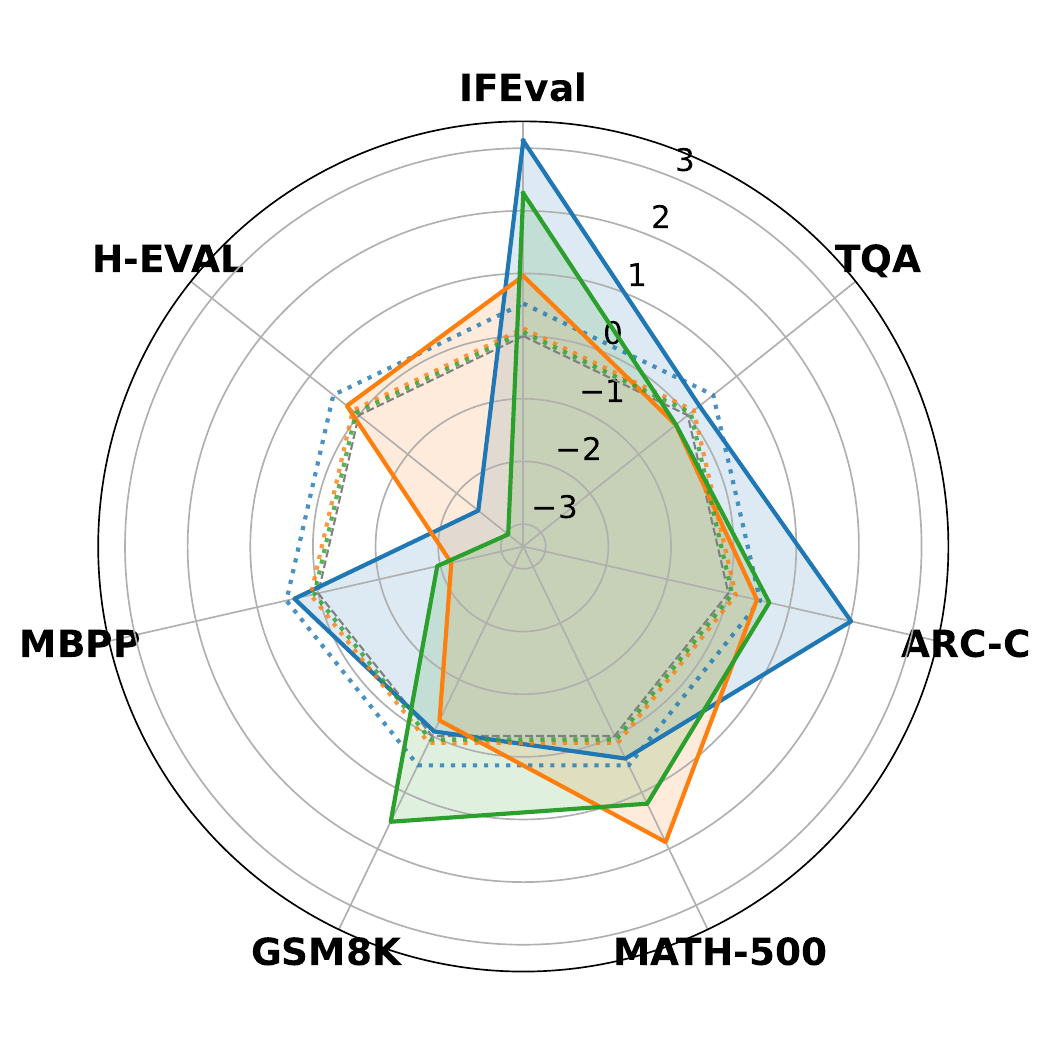}
    \caption{\mistralPlot}
    \label{subfig:mistral_7b_more_models}
  \end{subfigure}%

  \caption{\textbf{Using more models improves performance.} Relative score changes across benchmarks. Performance is compared between data generated using \textcolor{tabblue}{multiple models with judge}, \textcolor{taborange}{average of single pairs} and \textcolor{tabgreen}{multiple models with delta heuristic}, with dotted lines indicating average performance.}
  \label{fig:more_models_radar_plots}
\end{figure}

\paragraph{Results.} Figure~\ref{fig:more_models_radar_plots}  shows that, relative to the average single pair, increasing generator diversity improves the average score for three of the four trained models. The effect is largest for \qwenTwoSmall and \gemmaTwoTiny, where average gains grow from $+0.55\%$ to $+1.45\%$ and from $+0.82\%$ to $+2.00\%$, with IFEval gains of up to $+6.36\%$ and $+8.63\%$ respectively. For \qwenOneHalfSmall the average gain is smaller ($+0.15\%$ to $+0.54\%$), but IFEval improves from $-0.05\%$ to $+9.11\%$. For \mistral, the average does not improve ($+0.14\%$ vs. $+0.06\%$), although IFEval gains persist. The benefit is therefore not uniform across models, and the improvement remains well below that of the judge-supervised dataset. Still, the consistent gain on instruction following, together with the average gains for most models, is what \hyperref[prop:prop_pooling_generators]{Proposition 2} predicts for a pipeline that only changes generator diversity. See Table~\ref{tab:avg_more_few_pairs} for additional details.

\section{Conclusion}

In this work, we have conducted a thorough analysis of the impact of the data generation process on training with Direct Preference Optimization (DPO). We focus on two approaches: the standard and computationally expensive strategy of generating completions from multiple models and ranking them using an LLM-as-a-judge, and the recently proposed alternative of using two models from the same family and deterministically defining preferences based on their relative size.

Through an extensive theoretical and empirical analysis, we show that the two-model approach consistently underperforms, regardless of the model family or size used for either the generator models or the model being trained. We further show that this  degradation is consistent with systematic, generator-dependent biases introduced by the limited and homogeneous set of generators, which can dominate the preference signal and hinder useful learning; chosen and rejected completions from a single pair can, in some settings, be separated almost perfectly using stylistic features alone.

Finally, we directly test whether simply increasing generator diversity, without an external judge, can mitigate this effect. Since parameter size no longer predicts completion quality once generators span multiple families, we instead use coarse, benchmark-derived tiers to construct diverse pairs. This approach improves performance over the single-pair baseline for most models tested supporting the previous theoretical and empirical analysis. Taken together, our results indicate that generator diversity, is a critical factor behind the effectiveness of multi-model preference data.

\section*{Limitations}

Although our experiments cover a wide range of models for both data generation and training, a potential limitation is that the largest models considered for training have 8 billion parameters, consistent with the scale explored in the paper introducing the delta learning hypothesis~\citep{delta_learning_hyp}. Our theoretical results are agnostic to model scale, so we expect the underlying mechanism to hold at larger scales as well. A possible direction for future work is therefore to validate whether our findings continue to hold at larger scales.

\section*{Acknowledgments}
This work is part of the Villanova project, financed by IPICEI-CIS, Prog. n. SA. 102519 -- CUP B29J24000850005. 

\bibliography{bibliography}    

\appendix

\section{Empirical Analysis of Style Bias}
\label{app:style_bias_empirical}

To quantify style bias, we extract binary and count-based formatting features from each completion $x$. The feature extractor maps text completions to an 11-dimensional style representation vector $f(x) \in \mathbb{R}^{11}$, summarized in Table~\ref{tab:gsm8k_style_features}.

\begin{table*}[t]
\centering
\small
\caption{\textbf{GSM8K Style Features.} Description and detection criteria used to extract the 11-dimensional style representation vector.}
\label{tab:gsm8k_style_features}
\begin{tabular}{lll}
\toprule
\textbf{Feature Name} & \textbf{Type} & \textbf{Detection Criterion / Pattern} \\
\midrule
\texttt{has\_boxed\_latex} & Binary & Contains \texttt{\textbackslash boxed} or \texttt{\$\textbackslash boxed} \\
\texttt{has\_hash\_delimiter} & Binary & Contains \texttt{\#\#\#\#} (GSM8K answer marker) \\
\texttt{has\_calc\_annotations} & Binary & Matches calculator annotations \texttt{<<...>>} \\
\texttt{has\_display\_math\_dollars} & Binary & Contains display math delimiters \texttt{\$\$} \\
\texttt{has\_display\_math\_brackets} & Binary & Contains display math delimiters \texttt{\textbackslash [} and \texttt{\textbackslash ]} \\
\texttt{has\_bold\_answer} & Binary & Matches \texttt{**answer:**} prefix (case-insensitive) \\
\texttt{has\_plain\_answer\_prefix} & Binary & Matches unformatted \texttt{answer:} line prefix \\
\texttt{has\_bullet\_points} & Binary & Matches bulleted list items (\texttt{*} or \texttt{-}) \\
\texttt{has\_numbered\_list} & Binary & Matches numbered list items (e.g., \texttt{1.}) \\
\texttt{has\_bold\_text} & Binary & Contains Markdown bold syntax (\texttt{**}) \\
\texttt{latex\_symbol\_count} & Count & Total frequency of \texttt{\textbackslash} and \texttt{\$} characters \\
\bottomrule
\end{tabular}
\end{table*}

\paragraph{Classification Protocol.}
We train a Logistic Regression classifier on the extracted feature vectors to evaluate how linearly separable chosen and rejected completions are based purely on formatting. Models are evaluated using 5-fold cross-validation across all data splits grouping by prompt to avoid data leakage.

\section{Evaluation Details}

We use our own evaluation code, which for the benchmarks considered here is equivalent to \texttt{lm-evaluation-harness}~\citep{eval-harness}. Table~\ref{tab:benchmarks} reports the exact evaluation settings used for each benchmark to facilitate reproducibility.

\begin{table*}[htbp]
\centering
\caption{\textbf{Overview of evaluation benchmarks.}  All benchmarks are evaluated using a multi-turn chat format with Chain-of-Thought (CoT)~\citep{chainofthoughtpromptingelicitsreasoning} and few-shot examples injected as conversation turns where specified. The acronyms are used in the figures.}
\label{tab:benchmarks}
\begin{tabular}{lccccc}
\toprule
\textbf{Benchmark} & \textbf{CoT} & \textbf{Shots} & \textbf{Metric} & \textbf{Acronym} \\
\midrule
TruthfulQA & \ding{55} & 6 &   MC1 & TQA \\
Arc-Challenge & \ding{55} & 0 &  Unconditional Accuracy & ARC-C \\
IFEval & \ding{55} & 0 &  pass@1 (instance;loose) & - \\
IFBench & \ding{55} & 0 &  pass@1 (instance;loose) & - \\
Math-500 & \checkmark & 4 &   Flex EM & - \\
GSM8K & \checkmark & 8 &   EM & - \\
HumanEval & \ding{55} & 0 &  pass@1 & H-EVAL \\
MBPP & \ding{55} & 0 & pass@1 & - \\
\bottomrule
\end{tabular}
\end{table*}

\section{Additional Results}

This section provides the exact evaluation results for all experiments discussed in the main paper. Table~\ref{tab:model_evals_qwen4b_qwen0.6b} reports the preliminary results obtained with \qwenThreeSmall and \qwenThreeTiny. Table~\ref{tab:pairs_ablations} extends these findings by ablating the pairs used to generate the completions. Table~\ref{tab:olmo2_sft_dpo_ablation} then reports the results of experiments with \olmoTwoSft, which rule out the possibility of a data--model alignment issue. Finally, Table~\ref{tab:avg_more_few_pairs} provides a detailed comparison between the original dataset, the average obtained across individual pairs using the size heuristic, and the use of multiple models pairs determined through benchmarks. The latter setting is further divided into \fewPairs, which uses only the models considered in the previous experiments for data generation, and \morePairs, which additionally includes other models and correspond to the results discussed in Section~\ref{sec:improve_delta_learning}. The definitions of these two groups are provided in Appendix~\ref{sec:models_generator}.

\begin{table*}[htbp]
\centering
\caption{\textbf{Delta pair generation underperforms.} Benchmark deltas using the original \olmoDataset  dataset and \qwenThreeSmall and \qwenThreeTiny to generate completions.}
\label{tab:model_evals_qwen4b_qwen0.6b}
\resizebox{\textwidth}{!}{%
\begin{tabular}{l *{8}{c}}
\toprule
\textbf{Model / Dataset} & \textbf{IFEval} & \textbf{TQA} & \textbf{ARC-C} & \textbf{MATH-500} & \textbf{GSM8K} & \textbf{MBPP} & \textbf{H-EVAL} & \textbf{Avg} \\
\midrule
\row{\qwenTwoSmall}{40.17}{28.40}{45.48}{22.80}{53.37}{44.75}{46.34}
\rowColor{\olmoDataset}{9.23}{1.22}{2.68}{0.00}{3.64}{4.28}{1.22}
\rowColor{\olmoDatasetDeltaQwenSmall}{-0.36}{0.49}{0.67}{-0.60}{0.91}{1.17}{-1.22}
\midrule
\row{\gemmaTwoTiny}{67.87}{41.62}{48.49}{15.00}{54.51}{50.19}{40.24}
\rowColor{\olmoDataset}{12.11}{1.59}{1.34}{1.40}{0.91}{2.72}{6.71}
\rowColor{\olmoDatasetDeltaQwenSmall}{4.32}{1.22}{2.34}{-1.40}{2.27}{-1.56}{4.27}
\midrule
\row{\qwenOneHalfSmall}{40.53}{30.97}{44.48}{11.00}{48.52}{42.80}{33.54}
\rowColor{\olmoDataset}{13.67}{0.98}{2.01}{1.20}{1.67}{-3.50}{2.44}
\rowColor{\olmoDatasetDeltaQwenSmall}{-3.36}{0.37}{0.00}{1.60}{-0.53}{-1.56}{-3.66}
\midrule
\row{\qwenOneHalfBig}{52.88}{43.45}{41.47}{15.40}{60.50}{52.53}{45.12}
\rowColor{\olmoDataset}{5.64}{0.61}{-0.33}{-0.40}{0.15}{-1.95}{4.27}
\rowColor{\olmoDatasetDeltaQwenSmall}{-1.20}{0.37}{-2.34}{-2.60}{0.30}{-9.34}{-6.10}
\midrule
\row{\mistral}{64.75}{55.69}{52.51}{10.60}{45.03}{39.30}{39.63}
\rowColor{\olmoDataset}{3.12}{0.24}{2.01}{0.40}{-0.08}{0.39}{-2.44}
\rowColor{\olmoDatasetDeltaQwenSmall}{0.36}{-0.24}{-0.67}{2.00}{-0.99}{-5.45}{0.00}
\midrule
\row{\aya}{56.47}{31.70}{48.16}{6.60}{45.87}{42.02}{31.71}
\rowColor{\olmoDataset}{8.76}{3.06}{2.01}{0.20}{4.32}{-3.50}{-2.44}
\rowColor{\olmoDatasetDeltaQwenSmall}{0.12}{3.43}{1.00}{1.60}{5.08}{-4.67}{-1.22}
\bottomrule
\end{tabular}%
}
\end{table*}

\begin{table*}[htbp]
\centering
\caption{\textbf{Ablation pair generators show consistent low performance.} Benchmark deltas using the original \olmoDataset dataset and different model pairs to generate the completions.}
\label{tab:pairs_ablations}
\resizebox{\textwidth}{!}{%
\begin{tabular}{l *{8}{c}}
\toprule
\textbf{Model / Dataset} & \textbf{IFEval} & \textbf{TQA} & \textbf{ARC-C} & \textbf{MATH-500} & \textbf{GSM8K} & \textbf{MBPP} & \textbf{H-EVAL} & \textbf{Avg} \\
\midrule
\row{\qwenTwoSmall}{40.17}{28.40}{45.48}{22.80}{53.37}{44.75}{46.34}
\rowColor{\olmoDataset}
{9.23}{1.22}{2.68}{0.00}{3.64}{4.28}{1.22}
\cline{2-9} 
\row{\qwenThreeSmallTiny}{-0.36}{0.49}{0.67}{-0.60}{0.91}{1.17}{-1.22}
\row{\qwenThreeBigTiny}{0.36}{-0.12}{0.00}{0.00}{0.76}{-0.39}{0.61}
\row{\olmoTwoBigTiny}{5.76}{0.12}{1.00}{1.60}{-0.83}{1.56}{0.00}
\row{\olmoTwoSmallTiny}{4.32}{-0.24}{2.34}{1.00}{-0.76}{0.00}{-1.22}
\row{\gemmaTwoBigSmall}{1.44}{-0.12}{0.67}{0.40}{-0.23}{0.78}{-0.61}
\cline{2-9}
\rowColor{\avgPairs}{2.30}{0.02}{0.94}{0.48}{-0.03}{0.62}{-0.49}
\midrule

\row{\gemmaTwoTiny}{67.87}{41.62}{48.49}{15.00}{54.51}{50.19}{40.24}
\rowColor{\olmoDataset}{12.11}{1.59}{1.34}{1.40}{0.91}{2.72}{6.71}
\cline{2-9}
\row{\qwenThreeSmallTiny}{4.32}{1.22}{2.34}{-1.40}{2.27}{-1.56}{4.27}
\row{\qwenThreeBigTiny}{-0.84}{0.37}{0.00}{-0.60}{1.36}{1.95}{0.00}
\row{\olmoTwoBigTiny}{5.52}{0.12}{0.33}{-0.60}{1.36}{-2.33}{2.44}
\row{\olmoTwoSmallTiny}{0.84}{1.10}{0.33}{-1.80}{1.67}{-1.17}{3.05}
\row{\gemmaTwoBigSmall}{3.48}{0.24}{-0.67}{0.80}{0.76}{0.00}{-0.61}
\cline{2-9}
\rowColor{\avgPairs}{2.66}{0.61}{0.47}{-0.72}{1.49}{-0.62}{1.83}
\midrule

\row{\qwenOneHalfSmall}{40.53}{30.97}{44.48}{11.00}{48.52}{42.80}{33.54}
\rowColor{\olmoDataset}{13.67}{0.98}{2.01}{1.20}{1.67}{-3.50}{2.44}
\cline{2-9}
\row{\qwenThreeSmallTiny}{-3.36}{0.37}{0.00}{1.60}{-0.53}{-1.56}{-3.66}
\row{\qwenThreeBigTiny}{-0.60}{-0.24}{-0.33}{1.20}{0.38}{-0.39}{1.22}
\row{\olmoTwoBigTiny}{5.16}{-0.49}{1.00}{1.20}{-0.08}{-1.17}{1.22}
\row{\olmoTwoSmallTiny}{0.12}{-0.24}{-0.33}{-0.40}{0.15}{0.00}{2.44}
\row{\gemmaTwoBigSmall}{-1.56}{0.00}{0.00}{2.60}{-0.76}{-0.78}{3.05}
\cline{2-9}
\rowColor{\avgPairs}{-0.05}{-0.12}{0.07}{1.24}{-0.17}{-0.78}{0.85}
\midrule

\row{\mistral}{64.75}{55.69}{52.51}{10.60}{45.03}{39.30}{39.63}
\rowColor{\olmoDataset}{3.12}{0.24}{2.01}{0.40}{-0.08}{0.39}{-2.44}
\cline{2-9}
\row{\qwenThreeSmallTiny}{0.36}{-0.24}{-0.67}{2.00}{-0.99}{-5.45}{0.00}
\row{\qwenThreeBigTiny}{-0.12}{0.00}{0.00}{3.00}{0.00}{-3.89}{0.00}
\row{\olmoTwoBigTiny}{2.28}{-0.61}{1.34}{1.40}{0.45}{-1.17}{0.00}
\row{\olmoTwoSmallTiny}{1.80}{-0.37}{0.33}{1.80}{-0.23}{-1.17}{-0.61}
\row{\gemmaTwoBigSmall}{0.48}{0.00}{1.34}{1.20}{-0.61}{0.78}{2.44}
\cline{2-9}
\rowColor{\avgPairs}{0.96}{-0.24}{0.47}{1.88}{-0.27}{-2.18}{0.37}
\bottomrule
\end{tabular}%
}
\end{table*}

\begin{table*}[htbp]
\centering
\caption{\textbf{Delta pair doesn't underperform because of model-dataset alignment.} Benchmark deltas using the original \olmoDataset dataset and different model pairs to generate the completions training \olmoTwoSft, the model for which the dataset was designed.}
\label{tab:olmo2_sft_dpo_ablation}
\begin{tabular}{l *{4}{c} c}
\toprule
\textbf{Model / Dataset} & \textbf{IFEval} & \textbf{TQA} & \textbf{MATH-500} & \textbf{GSM8K} & \textbf{Avg} \\
\midrule
\olmoTwoSft & 61.63\%	& 26.93\%	& 14.60\%	&	51.86\%	&	38.75\% \\
\olmoDataset & \pos{7.79\%} & \pos{4.53\%} & \pos{0.60\%} & \pos{6.29\%} & \textbf{\pos{4.80\%}} \\
\cline{2-6} 
\qwenThreeSmallTiny & 0.24\% & 1.10\% & -0.80\% & 1.29\% & 0.46\% \\
\qwenThreeBigTiny & 1.08\% & 1.59\% & -1.20\% & 1.21\% & 0.67\% \\
\olmoTwoBigTiny & -0.48\% & -0.24\% & -0.60\% & -0.45\% & -0.44\% \\
\olmoTwoSmallTiny & -1.32\% & -0.37\% & -1.40\% & -0.45\% & -0.88\% \\
\gemmaTwoBigSmall & -1.08\% & -0.37\% & -1.60\% & 0.91\% & -0.53\% \\
\cline{2-6} 
\avgPairs & \negval{-0.31\%} & \pos{0.34\%} & \negval{-1.12\%} & \pos{0.50\%} & \textbf{\negval{-0.14\%}} \\
\bottomrule
\end{tabular}
\end{table*}

\begin{table*}[htbp]
\centering
\caption{\textbf{Using Multiple Pairs.} Benchmark deltas using the original \olmoDataset dataset and multiple pairs to generate the completions. \fewPairs and \morePairs correspond respectively to completions generated by only the models used in previous experiments and the ones listed in Appendix~\ref{sec:models_generator}}
\label{tab:avg_more_few_pairs}
\resizebox{\textwidth}{!}{%
\begin{tabular}{l *{9}{c}}
\toprule
\textbf{Model / Dataset} & \textbf{IFEval} & \textbf{TQA} & \textbf{ARC-C} & \textbf{MATH-500} & \textbf{GSM8K} & \textbf{MBPP} & \textbf{H-EVAL} & \textbf{Avg} \\
\midrule
\row{\qwenTwoSmall}{40.17}{28.40}{45.48}{22.80}{53.37}{44.75}{46.34}
\rowColor{\olmoDataset}{9.23}{1.22}{2.68}{0.00}{3.64}{4.28}{1.22}
\rowColor{\avgPairs}{2.30}{0.02}{0.94}{0.48}{-0.03}{0.62}{-0.49}
\rowColor{\fewPairs}{2.04}{-0.37}{0.33}{1.80}{-1.44}{1.95}{1.22}
\rowColor{\morePairs}{6.36}{-0.24}{0.33}{2.00}{0.76}{1.56}{-0.61}
\midrule
\row{\gemmaTwoTiny}{67.87}{41.62}{48.49}{15.00}{54.51}{50.19}{40.24}
\rowColor{\olmoDataset}{12.11}{1.59}{1.34}{1.40}{0.91}{2.72}{6.71}
\rowColor{\avgPairs}{2.66}{0.61}{0.47}{-0.72}{1.49}{-0.62}{1.83}
\rowColor{\fewPairs}{8.03}{0.37}{1.34}{-2.80}{-1.36}{-0.78}{3.66}
\rowColor{\morePairs}{8.63}{0.61}{-1.67}{-2.20}{0.23}{2.33}{6.10}
\midrule
\row{\qwenOneHalfSmall}{40.53}{30.97}{44.48}{11.00}{48.52}{42.80}{33.54}
\rowColor{\olmoDataset}{13.67}{0.98}{2.01}{1.20}{1.67}{-3.50}{2.44}
\rowColor{\avgPairs}{-0.05}{-0.12}{0.07}{1.24}{-0.17}{-0.78}{0.85}
\rowColor{\fewPairs}{0.12}{-0.24}{0.00}{1.60}{-0.08}{0.00}{-0.61}
\rowColor{\morePairs}{9.11}{-0.73}{1.00}{-0.40}{-3.11}{-2.72}{0.61}
\midrule
\row{\mistral}{64.75}{55.69}{52.51}{10.60}{45.03}{39.30}{39.63}
\rowColor{\olmoDataset}{3.12}{0.24}{2.01}{0.40}{-0.08}{0.39}{-2.44}
\rowColor{\avgPairs}{0.96}{-0.24}{0.47}{1.88}{-0.27}{-2.18}{0.37}
\rowColor{\fewPairs}{0.60}{-0.49}{0.00}{3.80}{0.00}{-2.33}{1.83}
\rowColor{\morePairs}{2.28}{-0.24}{0.67}{1.20}{1.52}{-1.95}{-3.05}
\bottomrule
\end{tabular}%
}
\end{table*}

\section{Data Generation}
\label{sec:data_gen}

This section provides a complete overview of data generation and explains how the datasets are constructed.

\subsection{Settings}
\label{sec:data_gen_settings}
We generate all data using vLLM~\citep{vllm} with the following parameters: \texttt{top\_p}=0.9, \texttt{temperature}=0.2, thinking disabled, and \texttt{max\_new\_tokens}=2048. We discard truncated completions to avoid introducing truncation bias. The same generation parameters are used across all models. All other generation parameters are set to their default values.

\subsection{Models Generator}
\label{sec:models_generator}
  
For the initial experiments in Section~\ref{sec:delta_learning_worse_than_llm_judge}, we use the models already introduced therein. Instead here we focus on the construction of preference pairs for the experiments in Section~\ref{sec:improve_delta_learning} (fully reported in Table~\ref{tab:avg_more_few_pairs}). 

For \fewPairs, chosen and rejected completions are sampled using the original model set. Specifically, following the parameter-size heuristic, chosen completions are sampled from $\{$\qwenThreeBig, \olmoTwoBig, \gemmaTwoBig$\}$ and rejected completions from $\{$\qwenThreeTiny, \olmoTwoTiny, \gemmaTwoSmall, \olmoTwoSmall$\}$. 

Then given the observed improvement, we enlarge the pool of models. However this is non-trivial, since as we add more models from different families the delta size heuristic fails to actually correlate with the quality of the completions. For example, as shown in Table~\ref{tab:model_evals_qwen4b_qwen0.6b}, \gemmaTwoTiny significantly outperforms \qwenOneHalfSmall despite having half the parameter count. For this reason, we use roughly the same models used when constructing the original \olmoDataset which differ greatly by their abilities and allow establishing deltas more easily. We evaluate models across Math, Instruction-Following, and Coding benchmarks as proxies for expected completion quality on \olmoDataset given its composition (see Table~\ref{tab:models_generator_eval} for benchmark scores).

Based on these capability tiers, we partition models into candidate groups $G_k = (C_k, R_k)$, where $C_k$ and $R_k$ denote the set of chosen and rejected generators, respectively:

\begin{enumerate}
    \item \textbf{Tier 1 vs.\ Tier 3 (Max Delta):} \\
    Chosen $C_1$: $\{$\gemmaTwoBig, \qwenTwoHalf, \gemmaTwoSmall$\}$; \\
    Rejected $R_1$: $\{$\tuluTwo, \qwenTwoTiny, \yi $\}$
    
    \item \textbf{Tier 1 vs.\ Tier 2 (High Quality, Subtle Delta):} \\
    Chosen $C_2$: $\{$\gemmaTwoBig, \qwenTwoHalf\}; \\
    Rejected $R_2$: $\{$\olmoTwoTiny, \mistral, \smolLMTwo$\}$
    
    \item \textbf{Tier 2 vs.\ Tier 3 (Moderate Delta):} \\
    Chosen $C_3$: $\{$\olmoTwoTiny, \mistral, \smolLMTwo$\}$; \\
    Rejected $R_3$: $\{$\tuluTwo, \qwenTwoTiny, \yi$\}$
\end{enumerate}

To generate a preference pair $(y_c, y_r)$ for a given prompt, we first uniformly sample a group $G_k \sim \{G_1, G_2, G_3\}$, and then uniformly sample individual generators $m_c \sim C_k$ and $m_r \sim R_k$. This process yields a dataset with a diversified set of generators which makes relying entirely on spurious features to learn the preference harder.

\begin{table*}[htbp]
\centering
\caption{\textbf{Models generator evaluation.} Performance across Math, Instruction-Following, and Code benchmarks to determine pairs group for DPO. For IFEval we report prompt-level strict accuracy, as a stricter metric widens the performance gap across models.}
\label{tab:models_generator_eval}
\resizebox{\textwidth}{!}{%
\begin{tabular}{l *{5}{c}}
\toprule
\textbf{Model} & \textbf{GSM8K} & \textbf{IFEval} & \textbf{MBPP} & \textbf{Avg} \\
\midrule 
\gemmaTwoBig & 80.67\% & 77.45\% & 78.60\% & \textbf{78.91\%} \\
\qwenTwoHalf & 83.70\% & 71.35\% & 73.93\% & \textbf{76.33\%} \\
\gemmaTwoSmall & 79.53\% & 73.01\% & 65.76\% & \textbf{72.77\%} \\
\olmoTwoTiny & 67.48\% & 63.96\% & 22.57\% & \textbf{51.33\%} \\ 
\smolLMTwo & 46.25\% & 46.95\% & 49.81\% & \textbf{47.67\%} \\
\mistral & 45.03\% & 48.98\% & 39.30\% & \textbf{44.44\%} \\
\tuluTwo & 31.92\% & 46.65\% & 40.86\% & \textbf{37.14\%} \\
\yi & 40.94\% & 26.06\% & 25.29\% & \textbf{30.76\%} \\
\qwenTwoTiny & 31.08\% & 16.27\% & 29.57\% & \textbf{25.64\%} \\
\bottomrule
\end{tabular}%
}
\end{table*}

\section{More Training Details}
\label{app:training_details}

We implement our training pipeline using the Hugging Face ecosystem, primarily relying on the \href{https://huggingface.co/docs/trl/en/index}{Transformers Reinforcement Learning (TRL)} library. We use DeepSpeed~\citep{DeepSpeed} and its implementation of ZeRO~\citep{ZeRO}, selecting the ZeRO stage according to model size. All experiments are conducted on a single node equipped with 8 H100 GPUs. We use \olmoDataset as the baseline for all experiments and we replace the original completions while keeping the underlying prompt distribution fixed. Our setup closely matches the training \href{https://github.com/allenai/open-instruct/blob/main/docs/olmo2.md}{configuration} of \olmoTwoSft, for which the baseline dataset was originally designed.

Across all experiments, we train for 1,000 optimization steps with an effective batch size of 128 and a maximum sequence length of 2,048 tokens. We use a linear learning-rate scheduler with a warmup ratio of $0.1$. For optimization, we use AdamW~\citep{adamw} with $\beta_1=0.9$, $\beta_2=0.999$, and $\epsilon=10^{-8}$. This fixed training budget and shared configuration provide a consistent basis for comparing performance across experimental settings.

We sweep the regularization parameter $\beta \in \{5, 10\}$ and adjust the learning-rate grid according to the dataset used. For the original \olmoDataset, we sweep learning rates in $\{1\times 10^{-6}, 5\times 10^{-7}, 1\times 10^{-7}\}$. In the initial experiments reported in Table~\ref{tab:model_evals_qwen4b_qwen0.6b} and some of the ablations in Table~\ref{tab:pairs_ablations}, we used the same grid for generated completions. However, we observed that a learning rate of $1\times 10^{-6}$ often caused substantial performance degradation, while the best results were frequently obtained with $1\times 10^{-7}$. We therefore expanded the search toward lower learning rates to enable a fairer comparison across data sources and avoid systematically disadvantaging generated completions. Specifically, for our generated completions, we use $\{5\times 10^{-7}, 1\times 10^{-7}, 5\times 10^{-8}, 1\times 10^{-8}\}$, as higher learning rates on pairs constructed using the delta-size heuristic consistently led to severe performance degradation, in some cases causing performance to fall below the SFT baseline. This sensitivity further supports the claim that training on a single model pair substantially alters the training dynamics, as discussed in Section~\ref{sec:train_dynamics}.

\end{document}